%% file: main.tex
\documentclass[10pt,twocolumn,letterpaper]{article}
\usepackage[T1]{fontenc}
\usepackage{xfp}
\usepackage[pagenumbers]{cvpr}
\usepackage{colortbl}
\usepackage{tabularx,array}
\usepackage{placeins}
\input{table-style}
\definecolor{cvprblue}{rgb}{0.21,0.49,0.74}
\usepackage[globalcitecopy]{bibunits}
\defaultbibliographystyle{ieeenat_fullname}
\defaultbibliography{refs1}
\usepackage[pagebackref,breaklinks,colorlinks,allcolors=cvprblue,hypertexnames=false]{hyperref}
\def\confName{CVPR}
\def\confYear{2027}
\title{CoverMem: Coverage-Constrained Visual Memory for Budgeted Embodied Question Answering}
\author{%
\begin{minipage}{0.97\textwidth}
\centering
\normalsize
Ruijie Ma\textsuperscript{1}\quad
Haoyue Liu\textsuperscript{1}\quad
Mingda Zhang\textsuperscript{2}\quad
Dongxu Yue\textsuperscript{3}\quad
Xiaoying Tang\textsuperscript{1,4,5,*}\\[4pt]
\small
\textsuperscript{1}School of Science and Engineering, The Chinese University of Hong Kong, Shenzhen\\
\textsuperscript{2}School of Artificial Intelligence, The Chinese University of Hong Kong\\
\textsuperscript{3}Tsinghua University, China\\
\textsuperscript{4}Shenzhen Future Network of Intelligence Institute (FNii-Shenzhen)\\
\textsuperscript{5}Guangdong Provincial Key Laboratory of Future Networks of Intelligence, CUHK(SZ)\\[4pt]
\footnotesize
\texttt{ruijiema@link.cuhk.edu.cn}\quad
\texttt{tangxiaoying@cuhk.edu.cn}\\[2pt]
\textsuperscript{*}Corresponding author.
\end{minipage}%
}
\begin{document}
\makeatletter
\g@addto@macro\@maketitle{%
  \begin{minipage}{\textwidth}
  \captionsetup{type=figure}
  \centering
  \includegraphics[width=0.75\linewidth]{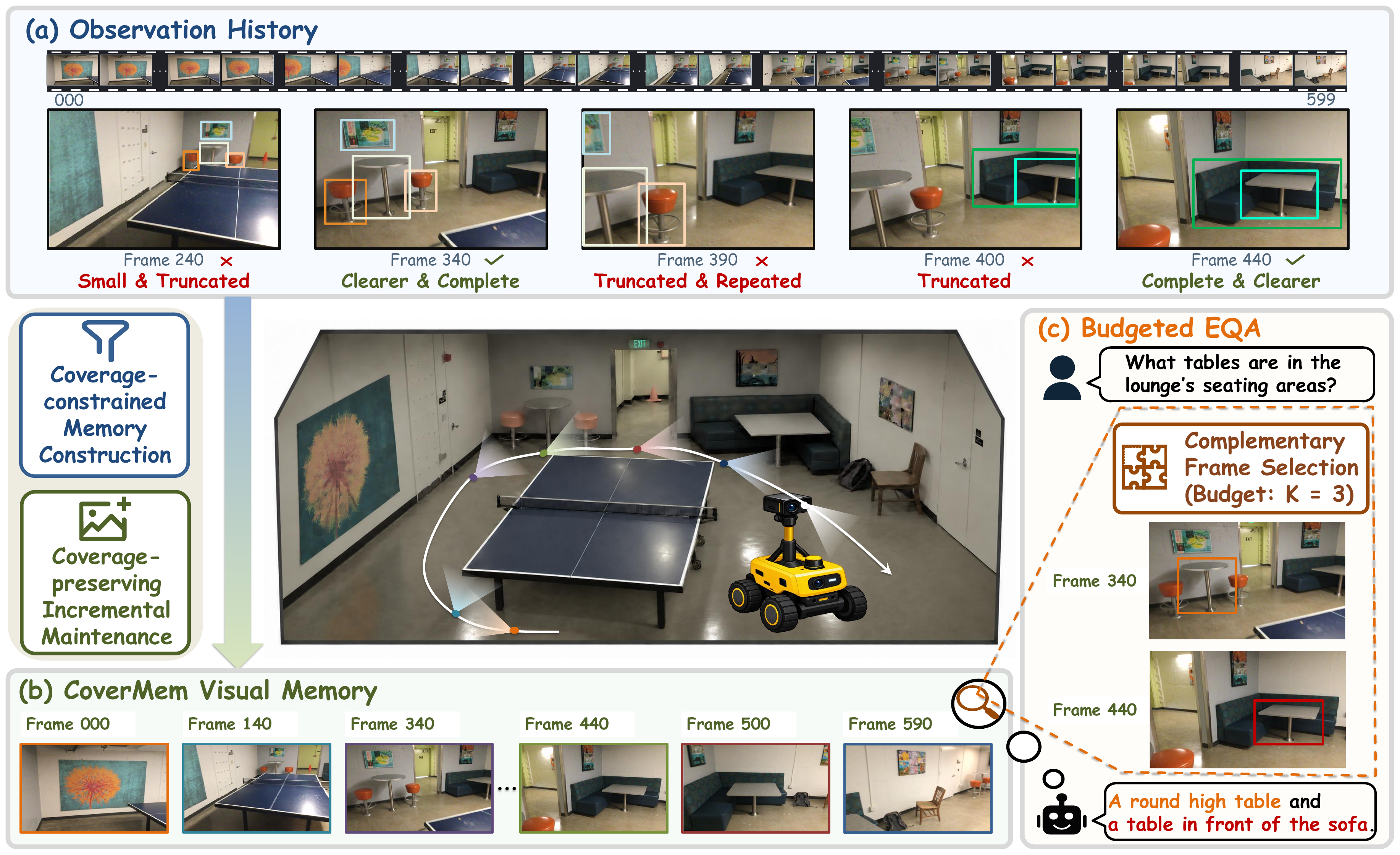}
  \caption{\textbf{Overview of CoverMem.} (a) Historical lounge observations differ in object visibility and completeness. (b) Reusable full-frame memory retains object details and scene context. (c) Complementary source frames provide evidence for budgeted question answering.}
  \label{fig:motivation}
  \end{minipage}\par\vspace{10pt}%
}
\makeatother
\maketitle
\flushbottom
\begingroup
\renewcommand{\endabstract}{\par}
\input{sections/00-abstract}
\endgroup
\clearpage
\input{sections/01-introduction}
\begin{figure*}[!t]
\centering
\includegraphics[width=\linewidth]{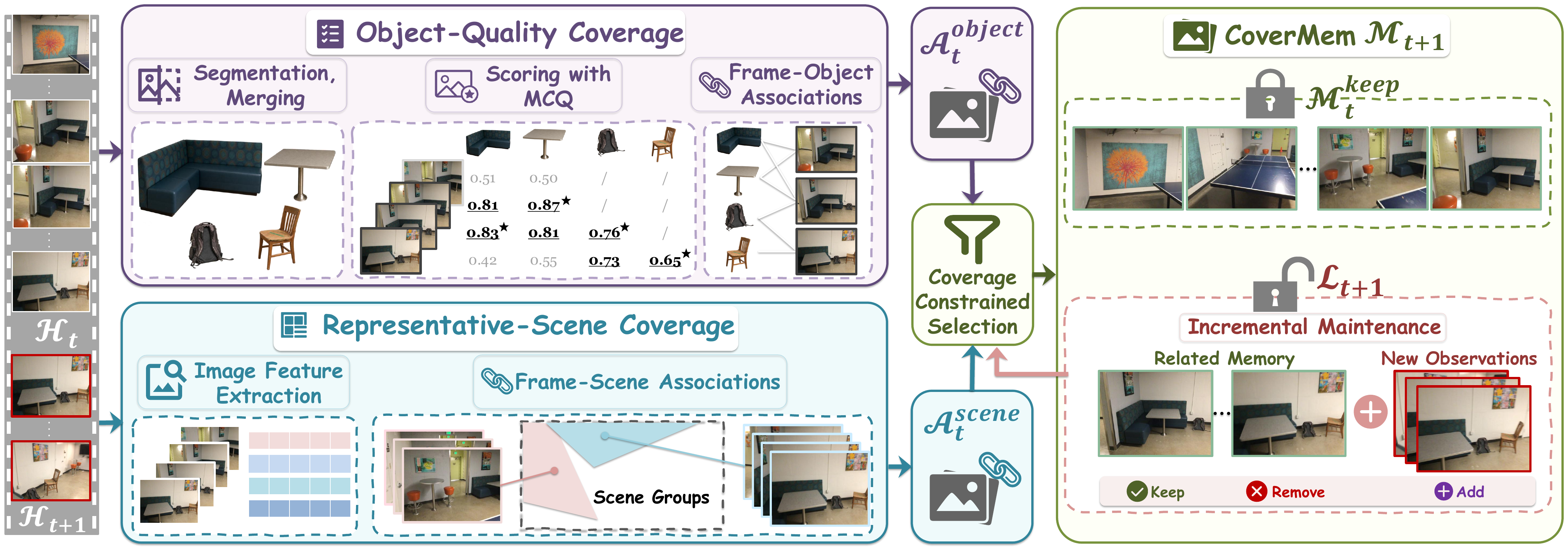}
\caption{\textbf{Memory construction and maintenance.}
Object-quality and representative-scene coverage guide compact memory construction.
Accumulated observations trigger local reselection of related
frames while unaffected frames remain fixed and coverage is preserved. Local candidates include new frames and memory frames sharing objects or representative scenes with them. Illustrative example.}
\label{fig:memory}
\end{figure*}

\input{sections/02-related}
\input{sections/03-method}
\input{sections/04-experiments}
\FloatBarrier
\input{sections/05-conclusion}
\par\pagebreak[4]
\clearpage
\flushbottom
\input{sections/07-llm-usage}
{\small
\bibliographystyle{ieeenat_fullname}
\bibliography{refs1}
}

\begin{bibunit}

\makeatletter

\let\CoverMemOriginalBackout\Hy@backout
\def\Hy@backout#1{%
  \begingroup
  \let\@auxout\@mainaux
  \CoverMemOriginalBackout{#1}%
  \endgroup
}

\def\hyper@natlinkstart#1{%
  \Hy@backout{#1}%
  \hyper@linkstart{cite}{cite.app.#1}%
  \def\hyper@nat@current{app.#1}%
}

\def\hyper@natlinkbreak#1#2{%
  \hyper@linkend#1%
  \hyper@linkstart{cite}{cite.app.#2}%
}

\def\hyper@natanchorstart#1{%
  \Hy@raisedlink{\hyper@anchorstart{cite.app.#1}}%
}

\makeatother

\input{sections/06-appendix}
\FloatBarrier
{\small
\putbib[refs1]
}
\end{bibunit}
\end{document}

%% file: table-style.tex
\newcolumntype{C}[1]{>{\centering\arraybackslash}m{#1}}
\newcolumntype{Y}{>{\centering\arraybackslash}X}
\renewcommand{\tabularxcolumn}[1]{m{#1}}
\usepackage{multirow}
\usepackage{trimclip}
\usepackage{makecell}
\usepackage{pifont}
\definecolor{oursbg}{RGB}{252,236,228} %
\definecolor{barorange}{RGB}{244,177,131}
\newcommand{\fst}[1]{\textbf{#1}}
\newcommand{\snd}[1]{#1}
\newcommand{\trd}[1]{#1}
\newcommand{\hp}[1]{\textbf{#1}}
\newcommand{\hb}[1]{\textbf{#1}}
\newcommand{\oursrow}{\rowcolor{oursbg}}
\newcommand{\up}{\,\(\uparrow\)}
\newcommand{\down}{\,\(\downarrow\)}
\newcommand{\cmark}{\ding{51}}
\newcommand{\gain}[1]{\,{\fontsize{6.2}{7}\selectfont(#1)}}
\newcommand{\tabsetup}{%
  \setlength{\aboverulesep}{0pt}\setlength{\belowrulesep}{0pt}%
  \setlength{\extrarowheight}{0.6pt}\renewcommand{\arraystretch}{1.12}%
  \renewcommand{\tabularxcolumn}[1]{m{##1}}}
\newcommand{\sharebar}[1]{%
  \makebox[1.15cm][l]{\textcolor{barorange}{\rule[-0.15ex]{\fpeval{#1*0.0275}cm}{1.35ex}}}%
  \makebox[0.8cm][r]{#1\%}}

%% file: sections/00-abstract.tex
\begin{abstract}
Episodic-memory embodied question answering enables agents to answer questions based on previously recorded visual observations of the surrounding scenes. However, growing observation histories, varying observation quality across viewpoints, and limited computational resources make it challenging to construct and maintain compact visual memories while retrieving sufficient evidence for question answering. To address these challenges, we introduce CoverMem, which incorporates three key components: (1) a coverage-constrained memory construction method that selects a compact set of informative full frames satisfying object-quality and representative-scene coverage requirements; (2) a coverage-preserving incremental maintenance strategy that locally reselects only the affected frames to satisfy updated coverage requirements as observations arrive; and (3) a budgeted complementary frame selection method that balances question relevance, additional support for question requirements, and redundancy with selected frames within a fixed budget.
Experimental results on three benchmarks covering diverse environments and question types demonstrate that CoverMem improves question-answering performance under limited visual budgets.
\par
\end{abstract}

%% file: sections/01-introduction.tex
\section{Introduction}
\label{sec:introduction}
\label{sec:intro}
\suppressfloats[t]

Vision-language models (VLMs)~\cite{hong20233d,huang2023embodied}
enable agents to answer questions about objects, locations,
states, and spatial relations using historical visual
observations~\cite{majumdar2024openeqa}.
Episodic-memory embodied question answering (EM-EQA) depends
both on model reasoning and on the ability to retain and
retrieve useful visual evidence from growing observation
histories, often by compressing observations and retrieving
a few relevant frames per question.
Yet memory construction must select observations before
future questions are known, whereas answering requires
sufficient evidence under a limited source-frame budget.
Consequently, excessive compression may discard critical
details, while retaining more observations does not ensure
effective use of the available answering budget.

Existing research addresses this challenge through visual
memory organization and evidence retrieval.
The former organizes observations using view-based memories,
object- or relation-centric structures, and reconstructed
3D representations
~\cite{gu2024conceptgraphs,lu2026gsmem,yang20253d,jatavallabhula2023conceptfusion,huang2023visual,peng2023openscene},
supported by aggregation and incremental updates
~\cite{ginting2025enter,gu2024conceptgraphs,yang20253d}.
The latter uses semantic similarity, object- or location-guided
search, and relevance-diversity selection
~\cite{huang2026memory,ong2025r,peng2026qca},
sometimes interleaving retrieval and reasoning
~\cite{anwar2025remembr,ginting2025enter}.
However, retaining an object's presence does not ensure
a sufficiently informative observation, since views differ
in sharpness, visibility, and detail.
Moreover, high-quality views of individual objects may lack
the layout or contextual information needed
to answer questions.
During memory updates, new observations may show one object
more clearly but miss other objects or scene context in
earlier frames, so direct replacement can discard useful
information.
At question time, relevance or visual diversity alone may
yield overlapping evidence for some question aspects,
leaving others underrepresented despite useful observations
remaining in memory.

Figure~\ref{fig:motivation} provides a lounge example
for discussing memory retention and question-time frame selection.
For ``What tables are in the lounge's seating areas?'',
high question-relevance scores may cause retrieval to favor
repeated views of the round high table (frame~340),
overlooking the table in front of the sofa (frame~440).
We consider two related but distinct design questions.
Before a question is known, memory construction must retain
informative object observations together with scene context,
and maintenance must avoid losing useful information
as observations arrive.
After the question is given, evidence selection must choose
an informative set of frames to answer the question accurately
within the available visual input budget.

We introduce \textbf{\textsc{CoverMem}}, a visual memory framework
that combines coverage preservation before questions are known
with complementary evidence selection within the available
input image budget.
Before questions are known, CoverMem constructs a compact memory that meets object-quality thresholds relative
to historical best observations while covering representative scenes.
As new observations arrive, CoverMem locally reselects affected
frames while keeping unaffected frames fixed, then verifies
that the memory satisfies both constraints.
For each question, CoverMem evaluates each candidate's additional
support for question-derived requirements relative to selected evidence, alongside relevance and visual redundancy.

Experiments on three QA benchmarks with two answering models
evaluate answer quality under limited source-frame budgets,
while component analyses examine memory quality,
maintenance cost, and frame selection.

Our contributions are summarized as follows:
\begin{itemize}
    \item We propose compact memory construction under
    object-quality and representative-scene coverage constraints,
    with per-object thresholds relative to historical best observations
    and at least one frame per representative scene.

    \item We enable incremental memory maintenance that locally
    reselects affected frames under updated constraints, keeps
    unaffected frames fixed, and verifies that the resulting
    memory satisfies all updated constraints.

    \item We develop complementary frame selection under a fixed
    image budget, combining question relevance, additional support
    for question-derived requirements relative to selected evidence,
    and visual redundancy.
\end{itemize}

%% file: sections/02-related.tex
\section{Related work}
\label{sec:related}

\noindent\textbf{Embodied question answering.}
EQA with episodic memory~\cite{majumdar2024openeqa}
answers questions by retrieving relevant stored views~\cite{majumdar2024openeqa,yang20253d}, while active EQA navigates and explores for question-relevant evidence~\cite{cheng2025efficienteqa,das2018embodied,ginting2025enter,gordon2018iqa,ren2024explore,saxena2024grapheqa,zhang2026fast}. We maintain full-frame episodic memory and select a limited number of source frames for answering.

\par\noindent\textbf{Memory construction and maintenance.}
Image memories retain detail and context~\cite{huang2026memory,yang20253d}, while object-centric structures support retrieval and relational reasoning but may omit appearance and context~\cite{ginting2025enter,gu2024conceptgraphs}. Reconstructed 3D geometry and appearance support spatial reasoning and novel-view rendering~\cite{kerbl20233d,kerr2023lerf,lu2026gsmem,yu2026splatreasoner}. Methods integrate structured and visual evidence for question answering~\cite{lu2026gsmem,saxena2024grapheqa}, while 3D-Mem updates memory incrementally~\cite{yang20253d}. CoverMem enforces object-quality and representative-scene coverage and uses local reselection for updates, fixing unaffected frames and verifying both~constraints.

\par\noindent\textbf{Evidence retrieval and selection.}
Before answering, OpenEQA samples frames uniformly~\cite{majumdar2024openeqa}, while question-conditioned selection uses semantic similarity and question cues~\cite{huang2026memory,ong2025r,wang2025videotree,yu2026splatreasoner} or balances relevance and diversity~\cite{kitamura2026informative}. Others interleave retrieval and reasoning~\cite{anwar2025remembr,fan2025embodied,sohn2026r4,wang2024videoagent}, with GraphPad editing scene graphs~\cite{ali2025graphpad}. To limit evidence overlap, CoverMem combines relevance, visual redundancy, and additional requirement support relative to selected frames under a source-frame budget.

%% file: sections/03-method.tex
\section{Method}
\label{sec:method}
\label{sec:method1}

From observation history \(\mathcal H_t\) collected by time \(t\),
CoverMem constructs reusable visual memory independently
of questions, with active memory
\(\mathcal M_t\subseteq\mathcal H_t\).
To support this process as new observations accumulate, each retained frame's associations with detected object
instances and representative scene groups enable coverage evaluation and the identification of related frames during
incremental maintenance.
Given a question \(q\), CoverMem selects at most \(K\)
distinct source frames \(S_q\subseteq\mathcal M_t\)
for answering.
We first define the coverage constraints in Section~\ref{sec:method2},
then describe incremental maintenance through local
reselection in Section~\ref{sec:method3},
and finally introduce budgeted complementary frame
selection in Section~\ref{sec:method4}.

\begin{figure*}[t]
\centering
\includegraphics[width=\linewidth,trim=0 5.5bp 0 5.5bp,clip]{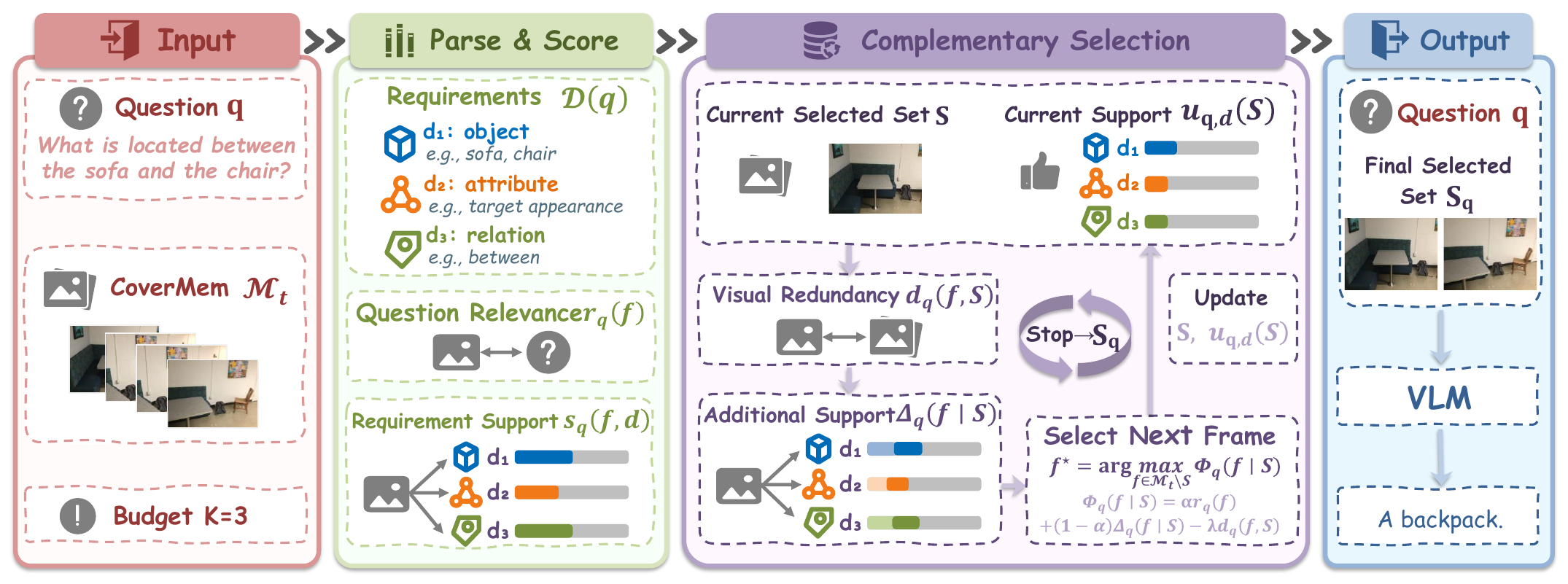}
\caption{\textbf{Budgeted complementary frame selection.}
Question-derived requirements guide source-frame selection.
Each step selects a frame using question relevance,
additional requirement support, and visual redundancy,
then updates the selected set and requirement support
until the budget is reached, candidates are exhausted,
or no score reaches the threshold. Illustrative example.}
\label{fig:access}
\end{figure*}

\subsection{Coverage-constrained memory construction}
\label{sec:method2}

As illustrated in Figure~\ref{fig:memory}, CoverMem constructs
compact memory under object-quality and representative-scene
coverage constraints. The former retain quality-qualified
object views, while the latter retain broader scene context.

\par\noindent\textbf{Object-quality coverage.}
Let \(\mathcal O_t\) denote the objects recorded up to time \(t\).
Coverage requires at least one quality-qualified
observation for each \(o\in\mathcal O_t\).
To compare observations linked across frames by object
associations, the multi-cue observation-quality (MCQ) score
\(Q(o,f)\) for object \(o\) in frame \(f\) combines six cues:
sharpness, object area, boundary truncation, detection
confidence, depth-based visibility, and text readability.
A weighted geometric mean aggregates these cues:
\begin{equation}
\label{eq:quality-aggregation}
Q(o,f)=\prod_{k=1}^{6}\tilde x_k(o,f)^{w_k}.
\end{equation}
Here, \(\tilde x_k(o,f)\in(0,1]\) are normalized cues,
with \(w_k\geq0\) and \(\sum_{k=1}^{6}w_k=1\).
Appendix~\ref{app:scoring} details cue definitions,
normalization, and weights.
For historical frames \(\mathcal H_t(o)\) linked to \(o\),
the object-specific quality threshold is set relative
to the object's best historical observation:
\begin{equation}
\label{eq:quality}
\theta_t(o)=\rho\max_{f\in\mathcal H_t(o)}Q(o,f),
\qquad \rho\in(0,1].
\end{equation}
With \(\rho\) fixed across objects, an associated frame
qualifies for \(o\) when \(Q(o,f)\geq\theta_t(o)\),
although it may not qualify for other objects.
Frame--object associations are given by
\(\mathcal A_t^{\mathrm{object}}
\subseteq \mathcal H_t\times\mathcal O_t\),
where \((f,o)\in\mathcal A_t^{\mathrm{object}}\)
indicates that object \(o\) is linked to frame \(f\).
An association contributes to object-quality coverage
only if the corresponding observation meets the quality
threshold for the associated object.

\par\noindent\textbf{Representative-scene coverage.}
Let \(\mathcal G_t\) denote the scene groups recorded up to time \(t\).
To complement object-level observations with broader context,
memory retains at least one frame associated with each
\(g\in\mathcal G_t\).
These groups are formed by visual similarity using global
image features, establishing the frame--scene connections
in Figure~\ref{fig:memory}.
Each group uses its first frame's feature as a fixed representative.
An incoming frame joins every group whose representative
matches its feature under a fixed cosine-similarity threshold,
or creates a new group if none matches.
Once established, these associations remain fixed, and groups may overlap.
The frame--scene associations are
\(\mathcal A_t^{\mathrm{scene}}
\subseteq \mathcal H_t\times\mathcal G_t\),
where \((f,g)\in\mathcal A_t^{\mathrm{scene}}\)
indicates that frame \(f\) is associated with the
representative scene group \(g\).

Together, the two association types determine which coverage
constraints each candidate frame can meet.
For selection, all historical frames are candidates,
\(\mathcal C_t=\mathcal H_t\).
Let \(\mathcal R_t\) collect the object-quality and
representative-scene coverage constraints as distinct types,
and let \(\mathcal R_t(f)\subseteq\mathcal R_t\) contain those
met by frame \(f\) under the quality and association
rules above.
The objective minimizes memory size subject to all constraints:
\begin{equation}
\label{eq:joint-coverage}
\min_{\mathcal M\subseteq\mathcal C_t} |\mathcal M|
\quad\text{s.t.}\quad
\bigcup_{f\in\mathcal M}\mathcal R_t(f)=\mathcal R_t.
\end{equation}
With the quality thresholds and scene associations fixed,
greedy set cover~\cite{chvatal1979greedy,wolsey1982analysis}
approximates this objective.
At each step, the frame meeting the most unmet
constraints is selected until all are met.
Redundant frames are removed only if the remaining
set meets every constraint.
The resulting memory meets all constraints, although
greedy selection does not guarantee a globally optimal memory.

\subsection{Coverage-preserving incremental \mbox{maintenance}}
\label{sec:method3}

Motivated by dynamic set cover~\cite{gupta2017online},
CoverMem maintains coverage through local reselection when
accumulated observations introduce new objects or scene groups,
or provide clearer views of previously recorded objects.

At update \(t+1\), let
\(\mathcal B_{t+1}=\mathcal H_{t+1}\setminus\mathcal H_t\)
contain frames collected since the previous update at \(t\).
Their object qualities and scene associations extend existing
records, updating historical bests without altering previous
scores or associations.
Improved historical bests raise observed objects' thresholds
and may disqualify earlier object--frame pairs, while thresholds
for absent objects remain unchanged.
To limit reselection cost, \(\mathcal L_{t+1}\) includes
\(\mathcal B_{t+1}\) and frames in \(\mathcal M_t\) sharing
an object or representative scene with any new frame,
using recorded associations before quality filtering.
The remaining frames stay fixed as
\(\mathcal M_t^{\mathrm{keep}}=
\mathcal M_t\setminus\mathcal L_{t+1}\).

Since affected frames may also cover other objects or scenes,
local reselection must meet the updated constraints left
unmet by the fixed memory:
\begin{equation}
\mathcal R_{t+1}^{\mathrm{res}}
=
\mathcal R_{t+1}
\setminus
\bigcup_{f\in\mathcal M_t^{\mathrm{keep}}}
\mathcal R_{t+1}(f).
\label{eq:residual-requirements}
\end{equation}
Greedy selection and redundancy removal from
Section~\ref{sec:method2} satisfy these residual constraints
within \(\mathcal L_{t+1}\).
The selected frames and \(\mathcal M_t^{\mathrm{keep}}\)
form \(\mathcal M_{t+1}\), whose coverage is verified against
all updated constraints for objects and scenes.

\subsection{Budgeted complementary frame selection}
\label{sec:method4}

At answering time, CoverMem selects complementary frames
within the image budget, as illustrated in Figure~\ref{fig:access}.
Selection combines relevance to the complete question,
additional support for question-derived visual requirements,
and visual redundancy with the frames already selected.

\par\noindent\textbf{Question-derived visual requirements.}
Each question \(q\) is first parsed into a nonempty set of
visual requirements \(\mathcal D(q)\), specifying the objects,
attributes, or relations needed for answering, with the
complete question used as a single requirement if parsing fails.
For each candidate frame \(f\in\mathcal M_t\),
relevance \(r_q(f)\in[0,1]\) matches the complete question
against global image features~\cite{radford2021learning},
while requirement support \(s_q(f,d)\in[0,1]\) combines
global image matching with the strongest regional evidence
from object crops and region descriptions, using global
matching alone when no regions are available
(Appendix~\ref{app:scoring}).

\par\noindent\textbf{Additional visual support.}
To assess evidence beyond selected frames, let \(S\)
denote the selected frame set, initialized as
\(S=\varnothing\).
Its current support for requirement \(d\) is
\(u_{q,d}(S)=\max_{g\in S}s_q(g,d)\),
with \(u_{q,d}(\varnothing)=0\).
A candidate's additional support is measured against this baseline:
\begin{equation}
\label{eq:support-gain}
\Delta_q(f\mid S)=
\frac{1}{|\mathcal D(q)|}
\sum_{d\in\mathcal D(q)}
\left[s_q(f,d)-u_{q,d}(S)\right]_+,
\end{equation}
where \([x]_+=\max(x,0)\) ensures that only improvements
over current requirement support contribute to the gain.

This gain differs from visual diversity because it measures
requirement support relative to selected evidence:
a visually different frame may add no support, whereas
a similar frame may reveal a needed detail more clearly.
To balance such gains with question relevance and visual
overlap, candidates are ranked by combining these three terms,
following the relevance--redundancy
principle~\cite{carbonell1998use}:
\begin{equation}
\label{eq:access}
\begin{aligned}
\Phi_q(f\mid S)
&=\alpha r_q(f)+(1-\alpha)\Delta_q(f\mid S)\\
&\quad-\lambda d_q(f,S),
\end{aligned}
\end{equation}
where \(\alpha\in[0,1]\), \(\lambda\geq0\),
and \(d_q(f,S)\) is the maximum normalized pairwise global
visual similarity between \(f\) and the currently selected
frames, with \(d_q(f,\varnothing)=0\).

At each step, CoverMem selects the highest-scoring
unselected frame until the budget is reached, candidates
are exhausted, or the best remaining score
\(\Phi_q(f\mid S)\) falls below the stopping threshold.
As \(S\) grows, additional support and redundancy change
while \(r_q(f)\) and \(s_q(f,d)\) remain fixed, allowing
candidate rankings to change after each selection.
Upon stopping, the selected full frames \(S_q\) and complete
question are passed to the answering model without feeding
intermediate answer reasoning back into selection.
The frame budget, however, does not guarantee sufficient
visual support for every question requirement at answering time.

%% file: sections/04-experiments.tex
\section{Experiments}
\label{sec:experiments}

We evaluate CoverMem through five research questions.
\textbf{Q1:} How does CoverMem compare with existing methods
across benchmarks and answering models? (\cref{sec:exp-a1})
\textbf{Q2:} How do coverage constraints and observation-quality
scoring affect memory quality and answering performance?
(\cref{sec:exp-a2})
\textbf{Q3:} Can local maintenance preserve coverage
while reducing maintenance cost? (\cref{sec:exp-a3})
\textbf{Q4:} How do selection components and candidate pools
affect answer quality? (\cref{sec:exp-a4})
\textbf{Q5:} How does the frame budget affect answer quality,
and what are the costs of memory construction and
question-time API usage? (\cref{sec:cost})

\input{tables/table1}

\subsection{Experimental setup}
\label{sec:setup}

\noindent\textbf{Datasets.}
Experiments use OpenEQA, EXPRESS, and
ScanQA~\cite{majumdar2024openeqa,jiang2025beyond,azuma2022scanqa}.
The fixed OpenEQA subset, denoted \emph{OpenEQA ScanNet},
contains 616 questions from 51 observation histories across
43 scenes.
EXPRESS contains 2,044 questions from 777 histories across
174 scenes, while the ScanQA validation subset contains
4,675 questions from 71 histories and scenes.
OpenEQA ScanNet and ScanQA share the ScanNet~\cite{dai2017scannet} visual source
but offer complementary evaluations through distinct
question--answer annotations.

\noindent\textbf{Metrics and protocol.}
We report LLM-Match for answer quality, the retained-frame
ratio (Ret. Ratio) for memory size, and the mean number
of distinct source frames supplied for answering (Avg. Frames).
For methods with explicit frame budgets, input is capped
at \(K=3\) source frames, while other methods retain their
input procedures.
Collage frames are counted before composition.
Memory analyses report the quality-qualified object ratio
(Qual. objects) and relative quality retention (Rel. quality),
measuring quality-threshold satisfaction and retained quality
relative to each object's best historical observation,
respectively.
Both metrics are computed per history and averaged equally
across histories, with definitions and implementation
details in Appendix~\ref{app:protocol}.

\noindent\textbf{Baselines.}
We compare with caption-based scene representations
(ConceptGraphs Captions~\cite{gu2024conceptgraphs}
and Sparse Voxel Map Captions~\cite{majumdar2024openeqa}),
visual memory and retrieval methods
(3D-Mem~\cite{yang20253d},
MemTree3D~\cite{huang2026memory},
VLM-Grounder~\cite{xu2024vlm},
QCA~\cite{peng2026qca},
WFS-SB~\cite{chen2026wavelet},
and ToolMerge~\cite{shlapentokh2026decomposing}),
and uniform-sampling and blind baselines.
All methods in the main comparison are rerun with
Qwen3.8-27B to generate answers, which GPT-6 Luna evaluates
under the same protocol.
CoverMem parses questions with Qwen3.5-35B-A3B and is
also evaluated with Gemini 3.5 Flash-Lite for answering.
Each answering model is paired with uniform sampling of
up to \(K\) distinct frames from the original history,
while the blind baseline receives only the question.

The following analyses of coverage constraints, memory
maintenance, frame selection, and computational cost use
the same fixed OpenEQA ScanNet evaluation subset.

\subsection{A1: Comparison with existing methods}
\label{sec:exp-a1}

\noindent\textbf{Comparison with existing methods.}
CoverMem achieves the highest LLM-Match scores among the
Qwen3.8-27B configurations on all three benchmarks
(Table~\ref{tab:table1}). Its advantage over caption-based
methods supports retaining visual evidence, while comparisons
with visual memory and retrieval methods highlight the
importance of evidence retention and selection.
CoverMem addresses these stages by preserving coverage before
questions are known and selecting complementary evidence
once a question arrives.
VLM-Grounder uses more frames but scores lower throughout.
CoverMem retains more frames than 3D-Mem on OpenEQA ScanNet
but fewer on EXPRESS and ScanQA, while scoring higher on
all three. These comparisons suggest that memory size alone
does not determine answer quality.
ToolMerge is the strongest competing Qwen3.8-27B baseline,
with CoverMem improving by 4.14, 1.03, and 6.66 points on
OpenEQA ScanNet, EXPRESS, and ScanQA, respectively.
The smaller margin on EXPRESS and larger margin on ScanQA
show that the gains vary across benchmarks under the same
answering model. All three improvements remain statistically
significant after Holm correction (adjusted \(p\leq0.03\);
Appendix~\ref{app:significance}).

\flushbottom
\noindent\textbf{Performance across answering models.}
CoverMem improves over paired uniform sampling with both
answering models using at most three source frames
(Table~\ref{tab:table1}). Since memory construction and
selection precede answer generation, selected evidence is
supplied through the same question-and-image interface.
Its consistent advantage across the evaluated model--benchmark
combinations suggests that the benefit of the evidence
pipeline extends beyond a single answering model.
Absolute scores reflect overall answer quality, whereas
paired gains measure improvement over each model's uniform
baseline. On EXPRESS, Gemini 3.5 Flash-Lite has a higher
absolute score but a smaller paired gain than Qwen3.8-27B,
owing to its stronger uniform baseline. Thus, higher absolute
performance does not imply a larger benefit from CoverMem.

\begin{figure}[!t]
\centering
\includegraphics[width=\linewidth]{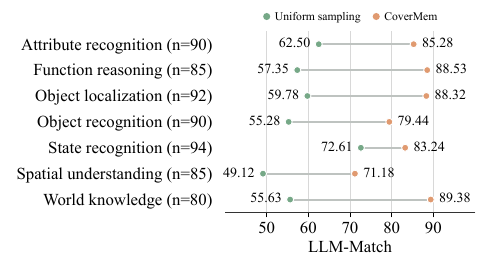}
\caption{\textbf{Performance across question types.}
Connected markers compare uniform sampling (green) and CoverMem
(orange). Labels show LLM-Match scores; \(n\) denotes question counts.}
\label{fig:question-types}
\end{figure}

\noindent\textbf{Performance across question types.}
CoverMem improves over uniform sampling in seven
categories under the same three-frame budget
(Figure~\ref{fig:question-types}).
The largest gains occur in world knowledge and function
reasoning, reaching 33.75 and 31.18 LLM-Match points,
respectively. These improvements indicate that CoverMem's
benefits extend across question types under a fixed input
budget. The contributions of memory retention and
complementary selection are examined separately in the
following ablations.
State recognition has the strongest uniform baseline at
72.61 but the smallest gain of 10.63 points.
Spatial understanding improves by 22.06 points yet remains
lowest at 71.18, suggesting that spatial relations remain
challenging even with retained object views and
representative scene context.

\subsection{A2: Coverage constraints and quality design}
\label{sec:exp-a2}

\input{tables/table2}
\input{tables/table3}

Replacing object-quality coverage with basic object coverage
reduces memory size but also lowers observation quality and
answer accuracy (Table~\ref{tab:table2}). The quality-qualified
object ratio falls from 100\% to 86\%, showing that retaining
an object's presence does not ensure an informative view. Representative-scene coverage contributes beyond
object-quality coverage. Without this constraint, the
quality-qualified object ratio remains at 100\%, yet
LLM-Match falls from 83.60 to 77.03 (Table~\ref{tab:table2}).
This variant achieves higher relative quality retention than removing
object-quality coverage but answers less accurately,
supporting broader scene context alongside object-level
evidence. However, meeting quality thresholds for recorded
objects does not ensure coverage of relevant unrecorded objects,
visible details, or spatial relations needed for future questions.

Since removing either constraint also changes memory size,
these ablations reflect quality--compactness trade-offs.
At matched per-history sizes, CoverMem outperforms Uniform
memory under the same downstream selection and answering
settings, supporting coverage-guided retention for observation
and answer quality. The 7.50-point LLM-Match gain at identical
per-history frame counts links the observed answer quality to the choice
of retained observations at a fixed memory size. This gain is significant after Holm
correction (adjusted \(p\leq0.03\);
Appendix~\ref{app:significance}).

\begin{figure}[!t]
\centering
\includegraphics[width=\linewidth,trim=0 14.16bp 0 5.28bp,clip]{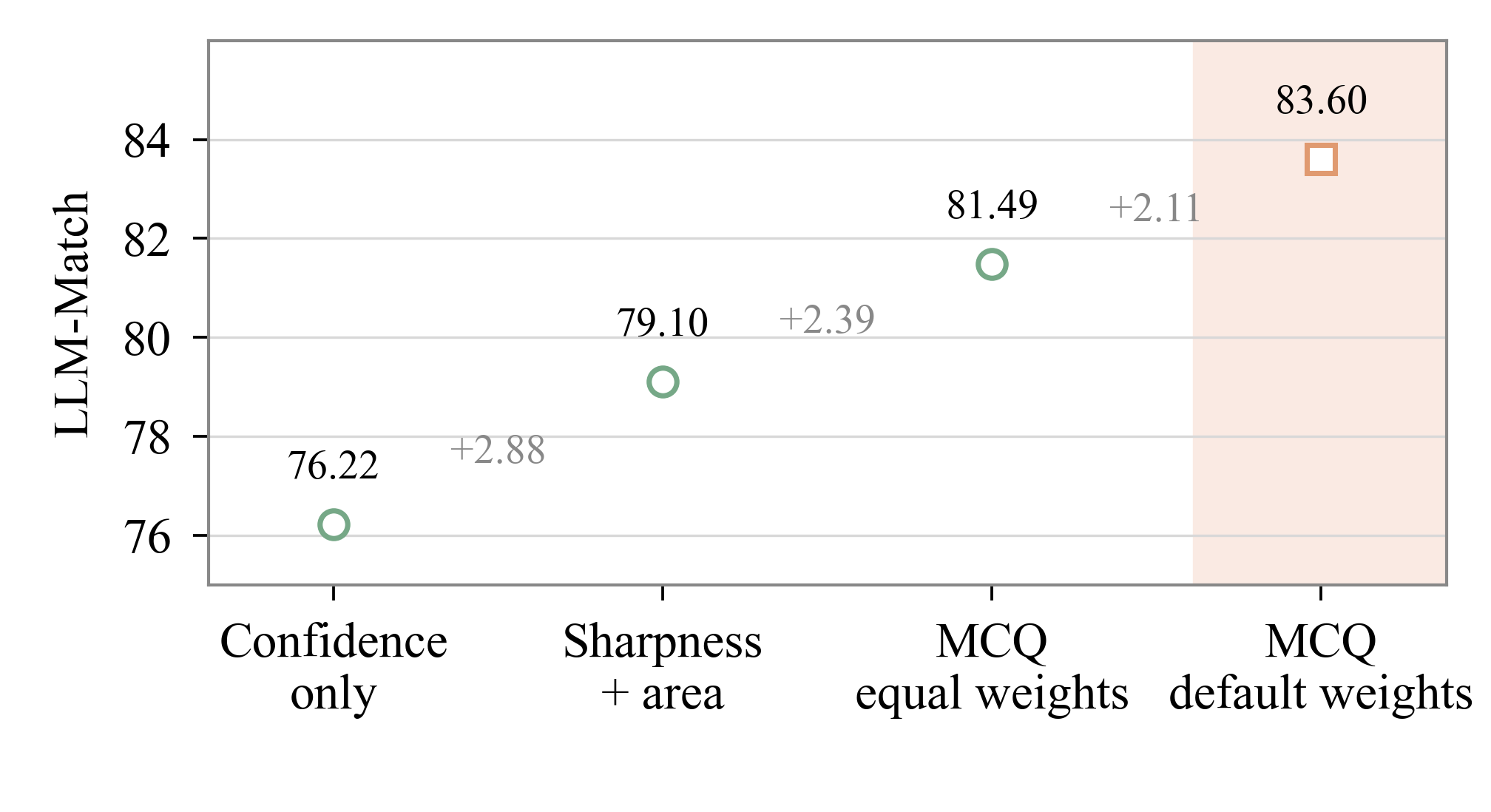}
\caption{\textbf{Observation-quality design.}
LLM-Match with at most \(K=3\) source frames per question.}
\label{fig:quality-design}
\end{figure}

With a source-frame budget of \(K=3\), we compare
quality-scoring strategies
(Figure~\ref{fig:quality-design}).
Confidence-only uses detector confidence as an observation-quality
proxy, while sharpness--area assesses visual clarity and object scale.
Equal-weight aggregation of all six cues improves over
sharpness--area by 2.39 LLM-Match points, and the default
weights add 2.11 points.
The default MCQ configuration achieves the highest
LLM-Match among the evaluated variants under the same
source-frame budget.

\input{tables/table4}

\subsection{A3: Analysis of incremental maintenance}
\label{sec:exp-a3}

Local maintenance and global reselection produce identical
final memory frame sets, per-question answers, and evaluation
scores on the evaluated histories (83.60 LLM-Match;
Table~\ref{tab:table3}). To quantify the update scope,
the reselection ratio is
\(r_{t+1}=|\mathcal L_{t+1}|/
|\mathcal M_t\cup\mathcal B_{t+1}|\),
where \(\mathcal L_{t+1}\) is the local candidate set from
Section~\ref{sec:method3}, reconsidered from \(t\) to
\(t+1\), and the denominator includes existing memory and
newly collected frames. Global reselection corresponds to
100\%. This ratio measures the local update scope relative
to current memory and newly collected frames, whereas
retention is measured against the original
observation history. Ratios are averaged within each history and then
equally across histories. By keeping unaffected frames fixed,
local maintenance reconsiders 49.30\% of the candidates and
reduces average maintenance time from 3.763 to 1.958 seconds
per history, a 48.0\% decrease. Local maintenance achieves
these savings while preserving memory frame sets and
answering outcomes on the evaluated histories. These timings exclude raw perception and API response waiting.

\subsection{A4: Complementary frame selection}
\label{sec:exp-a4}

Relevance-only evaluates frames against the complete question,
reaching 76.54 LLM-Match (Table~\ref{tab:table4}).
Adding requirement support provides a 4.43-point gain,
while adding the redundancy penalty yields 3.41 points.
Their combination performs best, with a 7.06-point gain.
These results favor combining additional support for question
requirements with visual diversity within the source-frame budget.

Requirement support, regional matching, and dynamic
updating are compared with memory, answering settings,
redundancy penalty, and source-frame budget fixed.
With relevance and redundancy retained, adding static global
support for parsed requirements raises LLM-Match
from 79.95 to 81.17. Adding regional matching increases
it to 81.86. With requirements and support scores fixed,
dynamic updating reaches 83.60 by assessing additional
support relative to selected evidence. The respective gains
of 1.22, 0.69, and 1.74 points favor all three steps,
with dynamic updating yielding the largest gain.

\begin{figure}[!t]
\centering
\includegraphics{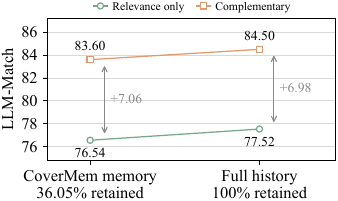}
\caption{\textbf{Frame selection across candidate pools.} OpenEQA ScanNet, Qwen3.8-27B, three-frame budget. For comparison, retention is measured relative to the original observation history.}
\label{fig:candidate-pools}
\end{figure}

Figure~\ref{fig:candidate-pools} compares two pipelines:
full history or compressed memory, followed by frame
selection and answering under the same model, evaluation
protocol, and source-frame budget. Complementary selection
improves LLM-Match over relevance-only selection by 6.98
and 7.06 points, respectively, favoring it within both pipelines.
Full history scores higher under either selection rule,
while compressed memory retains 63.95\% fewer frames
with score decreases below one point. Compressed memory
with complementary selection also outperforms the full-history
relevance-only pipeline by 6.08 points. More candidates
therefore do not necessarily yield better answers,
highlighting the trade-off between memory size and answer quality.

\input{tables/table6}
\input{tables/table7}

\subsection{A5: Budget sensitivity and cost}
\label{sec:cost}

\begin{figure}[!t]
\centering
\includegraphics{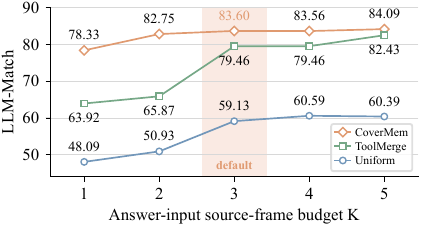}
\caption{\textbf{Source-frame budget.} LLM-Match on OpenEQA ScanNet for CoverMem, ToolMerge, and uniform sampling.}
\label{fig:budget}
\end{figure}

CoverMem outperforms uniform sampling and ToolMerge across
all evaluated source-frame budgets (Figure~\ref{fig:budget}).
Gains from additional frames diminish beyond three:
four frames provide no improvement, while five yield the
highest score, only 0.49 points above the three-frame result.
The margin over ToolMerge narrows from 4.14 points at three
frames to 1.66 at five, suggesting that additional inputs help
the baseline close the gap. Three frames therefore remain
a practical operating point for budgeted question answering.

Memory construction takes longer with CoverMem than with
the compared alternatives (Tables~\ref{tab:table6}
and~\ref{tab:table7}). This cost includes perception,
cross-frame association, quality estimation, and feature
processing needed to build reusable memory from observation
histories. Detection, segmentation, and geometric processing
form the largest component, accounting for 40.2\% of the
reported total construction time. At question time, CoverMem uses fewer parsing and answer-input
tokens than ToolMerge, with similar mean answer-output token
counts (Table~\ref{tab:table7}).
Uniform sampling requires no parsing or selection but uses
more answer-input tokens than CoverMem.
After parsing, candidate matching and complementary selection
run locally without requiring any further language-model API calls.

%% file: tables/table1.tex
\begin{table*}[t]
\centering
\caption{\textbf{Comparison across benchmarks and answering models.}
LLM-Match: answer quality; Ret. Ratio: retained-frame ratio;
Avg. Frames: mean number of answer-input frames.
\(\uparrow/\downarrow\): higher/lower is better;
``\textemdash{}'': not applicable;
\(100^{*}\): full candidate-pool retention.}
\label{tab:table1}
\fontsize{8}{9.6}\selectfont
\tabsetup
\setlength{\tabcolsep}{2pt}
\clipbox{0pt 0pt 0pt 0pt}{%
\begin{tabularx}{\linewidth}{@{}>{\raggedright\arraybackslash}m{0.165\linewidth}|*{2}{>{\hsize=1.3\hsize\linewidth=\hsize}Y>{\hsize=0.90\hsize\linewidth=\hsize}Y>{\hsize=0.80\hsize\linewidth=\hsize}Y|}>{\hsize=1.3\hsize\linewidth=\hsize}Y>{\hsize=0.90\hsize\linewidth=\hsize}Y>{\hsize=0.80\hsize\linewidth=\hsize}Y@{}}
\toprule
\multirow{2}{=}{\centering\textbf{Methods}}
 & \multicolumn{3}{c|}{\textbf{OpenEQA ScanNet}}
 & \multicolumn{3}{c|}{\textbf{EXPRESS}}
 & \multicolumn{3}{c}{\textbf{ScanQA}} \\
 & \hp{LLM-Match\up} & \hp{Ret. Ratio\newline(\%)\down} & \hp{Avg. Frames\down}
 & \hp{LLM-Match\up} & \hp{Ret. Ratio\newline(\%)\down} & \hp{Avg. Frames\down}
 & \hp{LLM-Match\up} & \hp{Ret. Ratio\newline(\%)\down} & \hp{Avg. Frames\down} \\
\midrule
\rowcolor{black!10}
\multicolumn{10}{c}{\textbf{Caption-based methods}} \\
ConceptGraphs~\cite{gu2024conceptgraphs}    & 36.97 & \textemdash{} & \textemdash{} & 36.61 & \textemdash{} & \textemdash{} & 39.78 & \textemdash{} & \textemdash{} \\
Sparse Voxel Map~\cite{majumdar2024openeqa} & 44.60 & \textemdash{} & \textemdash{} & 43.29 & \textemdash{} & \textemdash{} & 48.85 & \textemdash{} & \textemdash{} \\
\midrule
\rowcolor{black!10}
\multicolumn{10}{c}{\textbf{Memory methods and visual retrieval}} \\
3D-Mem~\cite{yang20253d}       & 62.05 & 27.54 & 2.68  & 65.58 & 33.51 & 3.00 & 57.53 & 43.75 & 2.84  \\
MemTree3D~\cite{huang2026memory}    & 64.65 & 100*   & 2.89  & 67.39 & 100*   & 2.84 & 65.71 & 100*   & 3.00  \\
VLM-Grounder~\cite{xu2024vlm} & 68.91 & 100*  & 20.09 & 69.84 & 100*  & 7.22 & 66.07 & 100*  & 15.29 \\
QCA~\cite{peng2026qca}          & \trd{69.56} & 100* & 3.00 & 73.67 & 100* & 3.00 & \trd{69.45} & 100* & 3.00 \\
WFS-SB~\cite{chen2026wavelet}       & 61.53 & 100* & 3.00 & \trd{76.35} & 100* & 3.00 & 66.87 & 100* & 3.00 \\
ToolMerge~\cite{shlapentokh2026decomposing}    & \snd{79.46} & 100* & 3.00 & \snd{77.96} & 100* & 3.00 & \snd{73.02} & 100* & 3.00 \\
\midrule
\rowcolor{black!10}
\multicolumn{10}{c}{\textbf{CoverMem and reference baselines}} \\
\multicolumn{10}{c}{Answering model: Qwen3.8-27B} \\
Blind (question only) & 36.61 & \textemdash{} & \textemdash{} & 45.66 & \textemdash{} & \textemdash{} & 36.76 & \textemdash{} & \textemdash{} \\
Uniform sampling & 59.13 & \textemdash{} & 3.00 & 61.57 & \textemdash{} & 3.00 & 50.22 & \textemdash{} & 3.00 \\
\oursrow
\textbf{CoverMem} & \fst{83.60}\gain{+24.47} & 36.05 & 2.62 & \fst{78.99}\gain{+17.42} & 24.94 & 2.51 & \fst{79.68}\gain{+29.46} & 22.40 & 3.00 \\
\multicolumn{10}{c}{Answering model: Gemini 3.5 Flash-Lite} \\
Uniform sampling & 62.30 & \textemdash{} & 3.00 & 72.35 & \textemdash{} & 3.00 & 53.67 & \textemdash{} & 3.00 \\
\oursrow
\textbf{CoverMem} & \textbf{87.38}\gain{+25.08} & 36.05 & 2.62 & \textbf{79.68}\gain{+7.33} & 24.94 & 2.51 & \textbf{78.99}\gain{+25.32} & 22.40 & 3.00 \\
\bottomrule
\end{tabularx}%
}

\end{table*}

%% file: tables/table2.tex
\begin{table}[!t]
\centering
\caption{\textbf{Memory construction ablations.} Uniform memory matches per-history size and uses CoverMem's selection.}
\label{tab:table2}
\fontsize{8}{9.6}\selectfont
\tabsetup
\setlength{\extrarowheight}{0pt}\renewcommand{\arraystretch}{1.0}
\setlength{\tabcolsep}{2.2pt}
\clipbox{0pt 0pt 0pt 0pt}{%
\begin{tabularx}{\linewidth}{@{}>{\raggedright\arraybackslash}m{0.24\linewidth}>{\hsize=0.89\hsize\linewidth=\hsize}Y>{\hsize=1.13\hsize\linewidth=\hsize}Y>{\hsize=1.03\hsize\linewidth=\hsize}Y>{\hsize=0.95\hsize\linewidth=\hsize}Y@{}}
\toprule
\textbf{Configuration} & \textbf{\makecell{Ret. Ratio\\(\%)\down}} & \textbf{\makecell{Qual. objects\\(\%)\up}} & \textbf{\makecell{Rel. quality\\(\%)\up}} & \textbf{LLM-Match\up} \\
\midrule
\rule[-3.4pt]{0pt}{11.4pt}w/o quality & \rule[-3.4pt]{0pt}{11.4pt}26.80 & \rule[-3.4pt]{0pt}{11.4pt}86.00 & \rule[-3.4pt]{0pt}{11.4pt}90.12 & \rule[-3.4pt]{0pt}{11.4pt}80.24 \\
\rule[-3.4pt]{0pt}{11.4pt}w/o scene & \rule[-3.4pt]{0pt}{11.4pt}30.56 & \rule[-3.4pt]{0pt}{11.4pt}100.00 & \rule[-3.4pt]{0pt}{11.4pt}97.31 & \rule[-3.4pt]{0pt}{11.4pt}77.03 \\
\rule[-3.4pt]{0pt}{11.4pt}Uniform memory & \rule[-3.4pt]{0pt}{11.4pt}36.05 & \rule[-3.4pt]{0pt}{11.4pt}83.85 & \rule[-3.4pt]{0pt}{11.4pt}88.97 & \rule[-3.4pt]{0pt}{11.4pt}76.10 \\
\oursrow
\rule[-3.4pt]{0pt}{11.4pt}\textbf{CoverMem} & \rule[-3.4pt]{0pt}{11.4pt}36.05 & \rule[-3.4pt]{0pt}{11.4pt}100.00 & \rule[-3.4pt]{0pt}{11.4pt}\textbf{98.49} & \rule[-3.4pt]{0pt}{11.4pt}\textbf{83.60} \\
\bottomrule
\end{tabularx}%
}
\end{table}

%% file: tables/table3.tex
\begin{table}[!t]
\centering
\caption{\textbf{Memory maintenance.} Means over 51 cached histories.}
\label{tab:table3}
\fontsize{8}{9.6}\selectfont
\tabsetup
\setlength{\extrarowheight}{0pt}\renewcommand{\arraystretch}{1.0}
\setlength{\tabcolsep}{3pt}
\clipbox{0pt 0pt 0pt 0pt}{%
\begin{tabularx}{\linewidth}{@{}>{\raggedright\arraybackslash}m{0.22\linewidth}YY@{}}
\toprule
\textbf{Maintenance} & \textbf{\makecell{Reselection ratio\\(\%)\down}} & \textbf{\makecell{Time / history\\(s)\down}} \\
\midrule
\rule[-3.4pt]{0pt}{11.4pt}Global & \rule[-3.4pt]{0pt}{11.4pt}100.00 & \rule[-3.4pt]{0pt}{11.4pt}3.763 \\
\oursrow
\rule[-3.4pt]{0pt}{11.4pt}\textbf{Local} & \rule[-3.4pt]{0pt}{11.4pt}\textbf{49.30} & \rule[-3.4pt]{0pt}{11.4pt}\textbf{1.958} \\
\bottomrule
\end{tabularx}%
}
\end{table}

%% file: tables/table4.tex
\begin{table}[!t]
\centering
\caption{\textbf{Ablation of frame selection.}
Support: requirement support, dynamic unless marked Static.
Regions: regional evidence added to global matching.
Redundancy: visual redundancy penalty.
Relevance: complete-question relevance.}
\label{tab:table4}
\fontsize{8}{9.6}\selectfont
\tabsetup
\setlength{\extrarowheight}{0pt}
\renewcommand{\arraystretch}{1.0}
\setlength{\tabcolsep}{1.5pt}
\begin{tabularx}{\linewidth}{YYYY|c}
\toprule
\multicolumn{4}{c|}{\textbf{Selection Configuration}}
& \multirow{2}{*}{\textbf{LLM-Match\up}} \\
\hb{Support} & \hb{Regions} & \hb{Redundancy}
& \hb{Relevance} & \\
\midrule
\rule[-3.4pt]{0pt}{11.4pt}
& & & \cmark & 76.54 \\
\rule[-3.4pt]{0pt}{11.4pt}\cmark
& \cmark & & \cmark & 80.97 \\
\rule[-3.4pt]{0pt}{11.4pt}
& & \cmark & \cmark & 79.95 \\
\rule[-3.4pt]{0pt}{11.4pt}Static
& & \cmark & \cmark & 81.17 \\
\rule[-3.4pt]{0pt}{11.4pt}Static
& \cmark & \cmark & \cmark & 81.86 \\
\oursrow
\rule[-3.4pt]{0pt}{11.4pt}\cmark
& \cmark & \cmark & \cmark & \fst{83.60} \\
\bottomrule
\end{tabularx}
\end{table}

%% file: tables/table6.tex
\begin{table}[!t]
\centering
\caption{\textbf{Memory construction time per history.}}
\label{tab:table6}
\fontsize{8}{9.6}\selectfont
\tabsetup
\setlength{\extrarowheight}{0pt}\renewcommand{\arraystretch}{1.0}
\setlength{\tabcolsep}{2.4pt}
\begin{tabularx}{\linewidth}{@{}>{\raggedright\arraybackslash}m{0.53\linewidth}|C{0.15\linewidth}|Y@{}}
\toprule
\textbf{Construction Component} & \textbf{Time (s)} & \textbf{Share of Total} \\
\midrule
Object detection, segmentation, and geometric processing & 17.80 & \sharebar{40.2} \\
Mask generation for region descriptions & 8.78 & \sharebar{19.8} \\
Region description generation & 6.99 & \sharebar{15.8} \\
Feature extraction, cross-frame association, quality computation, and related operations & 10.72 & \sharebar{24.2} \\
\midrule
\textbf{Total of listed components} & \textbf{44.29} & \makebox[1.15cm][l]{}\makebox[0.8cm][r]{\textbf{100\%}} \\
\bottomrule
\end{tabularx}
\end{table}

%% file: tables/table7.tex
\begin{table}[!t]
\centering
\caption{\textbf{Construction time and API tokens.} Time: cumulative s/history. Tokens: rounded means/question. Parsing includes ToolMerge planning. ``\textemdash{}'': no call.}
\label{tab:table7}
\fontsize{8}{9.6}\selectfont
\tabsetup
\setlength{\extrarowheight}{0pt}\renewcommand{\arraystretch}{1.0}
\setlength{\tabcolsep}{2pt}
\begin{tabularx}{\linewidth}{@{}>{\raggedright\arraybackslash}m{0.22\linewidth}>{\raggedright\arraybackslash}m{0.15\linewidth}|YY>{\columncolor{oursbg}}Y}
\toprule
\multicolumn{2}{@{}l|}{\textbf{Metric}} & \hb{Uniform} & \hb{ToolMerge} & \cellcolor{oursbg}\textbf{CoverMem} \\
\midrule
\multicolumn{2}{@{}l|}{\textbf{Construction time (s)\down}} & 0.46 & 33.74 & 44.29 \\
\midrule
\multicolumn{2}{@{}l|}{\textbf{API tokens per question}} & & & \cellcolor{oursbg} \\
\multirow{2}{*}{\textbf{Parsing}} & \hp{Input} & \textemdash{} & 1,732 & 415 \\
 & \hp{Output} & \textemdash{} & 182 & 35 \\
\multirow{2}{*}{\textbf{Answer}} & \hp{Input} & 2,545 & 2,545 & 2,184 \\
 & \hp{Output} & 99 & 100 & 101 \\
\multirow{2}{*}{\textbf{Evaluation}} & \hp{Input} & 238 & 237 & 233 \\
 & \hp{Output} & 44 & 49 & 41 \\
\bottomrule
\end{tabularx}
\end{table}

%% file: sections/05-conclusion.tex
\section{Conclusion}
CoverMem constructs compact full-frame memory for budgeted
embodied question answering under object-quality and
representative-scene coverage constraints.
It preserves relatively high-quality object observations
and scene context before receiving questions.
Incremental maintenance locally reselects affected frames
as observations arrive, preserving coverage.
Complementary frame selection combines question relevance,
additional requirement support, and visual redundancy
for question-specific evidence under a fixed budget.
Experiments on three benchmarks with two answering models
show improved performance using at most three source frames;
component analyses show high relative quality retention
and reduced maintenance time.
\label{lastmainpage}

%% file: sections/07-llm-usage.tex
\section*{LLM Usage Statement}
OpenAI's ChatGPT (GPT) was used solely for language polishing of the manuscript, including grammar, wording, and readability. The authors reviewed all suggested text and take full responsibility for the research design, implementation, data analysis, figures, experimental results, and all scientific claims presented in the paper.

%% file: sections/06-appendix.tex
\clearpage
\appendix
\maketitlesupplementary
\noindent{This supplement details the evaluation protocol, implementation, component analyses, streaming behavior, mathematical properties, and qualitative results of CoverMem, distinguishing reusable memory construction before a question is known from evidence allocation after the question is available.}

\FloatBarrier
\section{Evaluation and implementation details}\label{app:A}
\setcounter{table}{0}\renewcommand{\thetable}{A\arabic{table}}
\setcounter{figure}{0}\renewcommand{\thefigure}{A\arabic{figure}}

\subsection{Evaluation protocol and metrics}\label{app:protocol}

All comparisons use the fixed subsets described in the main paper. OpenEQA ScanNet and ScanQA share ScanNet~\cite{dai2017scannet} visual observations but use different question--answer annotations, providing complementary evaluations.

\begin{table}[t]\centering
\caption{Fixed evaluation subsets.}\label{tab:app-data}
\small\setlength{\tabcolsep}{4pt}\renewcommand{\arraystretch}{1.12}
\begin{tabularx}{\linewidth}{@{}>{\hsize=1.73585\hsize\linewidth=\hsize\raggedright\arraybackslash}X>{\hsize=0.75472\hsize\linewidth=\hsize\raggedright\arraybackslash}X>{\hsize=0.75472\hsize\linewidth=\hsize\raggedright\arraybackslash}X>{\hsize=0.75472\hsize\linewidth=\hsize\raggedright\arraybackslash}X@{}}
\toprule
\textbf{Benchmark} & \textbf{Questions} & \textbf{Histories} & \textbf{Scenes} \\
\midrule
OpenEQA ScanNet & 616 & 51 & 43 \\
EXPRESS & 2044 & 777 & 174 \\
ScanQA & 4675 & 71 & 71 \\
\bottomrule
\end{tabularx}
\end{table}

\noindent\textbf{Models and inputs.} The main comparison uses Qwen3.8-27B for answering, Qwen3.5-35B-A3B for question parsing, and GPT-6 Luna for evaluation. The additional answering-model comparison pairs CoverMem with uniform sampling using Gemini 3.5 Flash-Lite; external methods are compared under the main answering model. The default budget is at most $K=3$ distinct source frames.

\noindent\textbf{Answer evaluation.} The evaluator receives the question, reference answer, and predicted answer. We use the OpenEQA LLM-Match rating and conversion~\cite{majumdar2024openeqa} with the evaluator above, selecting the corresponding template when additional valid answers are available. For $N$ questions with integer ratings $r_i\in\{1,2,3,4,5\}$,

\[
\mathrm{LLM\text{-}Match}=\frac{25}{N}\sum_{i=1}^{N}(r_i-1).
\]

Scores average questions rather than equally weighting question categories. ScanQA results here use LLM-Match rather than native EM@1. Technical failures are recorded separately; main-paper comparisons use the same fixed question sets without technical exclusions. Evaluator choice affects the resulting scores. The independent automatic semantic and evidence assessments below provide diagnostic measurements, rather than human ground-truth annotations.

The main-model comparisons are reported in main-paper Table~\ref*{tab:table1}.

\noindent\textbf{Memory metrics.} Retained-frame ratio divides the retained source-frame count by the original historical count within the specified scope. Mean input frames average the number of distinct source frames supplied per question. Let $Q_h^*(o)$ and $Q_h^M(o)$ be the best historical and retained quality of object $o$ in history $h$, with retained quality zero when no observation remains. Both object metrics use equal weighting of histories after averaging objects within each history:

\[
\mathrm{Qualified}(M)=\frac{100}{|\mathcal H|}\sum_{h\in\mathcal H}\frac{n_h^{\mathrm{qual}}}{|\mathcal O_h|}.
\]

Here $n_h^{\mathrm{qual}}$ counts objects satisfying $Q_h^M(o)\geq0.7Q_h^*(o)$. Relative quality is

\[
\mathrm{RelQuality}(M)=\frac{100}{|\mathcal H|}\sum_{h\in\mathcal H}\frac{1}{|\mathcal O_h|}\sum_{o\in\mathcal O_h}\frac{Q_h^M(o)}{Q_h^*(o)}.
\]

Positive cue floors keep the historical reference positive for recorded objects. All cue-removal rows use this same complete six-cue reference and the fixed 0.7 qualification threshold for evaluation.

\subsection{CoverMem implementation}\label{app:scoring}

\noindent\textbf{Representations.} Full images and rectangular object crops use OpenCLIP ViT-L/14 with its OpenAI checkpoint. Region descriptions use sentence-t5-xl. These representations are 768-dimensional and unit-normalized. Questions and requirement search phrases use the corresponding CLIP text encoder for image matching and sentence-t5-xl for description matching. Raw scores are cosine similarities.

Crop coordinates are integer-clamped to image bounds. DAM-3B generates question-independent region descriptions from the full RGB image and binary region mask using focal prompting. Descriptions are encoded in the template \texttt{an object that looks like <description>}. Crops and descriptions serve retrieval, while the answering model receives the selected full images.

\noindent\textbf{Association interface.} Following 3D-Mem~\cite{yang20253d} and ConceptGraphs~\cite{gu2024conceptgraphs}, we use YOLO-World (\texttt{yolov8x-world.pt}) with the ScanNet200 vocabulary for object detection and SAM ViT-L (\texttt{sam\_l.pt}) for box-prompted instance segmentation. Object regions are projected into world coordinates using provided
depth maps, camera intrinsics, and poses. Cross-frame association combines 3D point-cloud overlap and CLIP ViT-B/32 appearance cosine similarity. Observations
with a combined matching score above 1.2 are merged into existing instances, with point clouds fused and appearance features updated; otherwise, new instances are created. Periodic deduplication further merges redundant instances. No ground-truth instance IDs or separate video trackers are used. Depth and camera parameters are provided by the dataset or simulator. Coverage is defined over the resulting recorded object identifiers rather than all physical instances. Within each benchmark, CoverMem ablations share observations and preprocessed object associations, except in the explicit association-perturbation tests. The streaming evaluation operates on cached association records, with previously established associations held fixed during memory replay.

\noindent\textbf{Question parsing.} The parser receives question text without images or reference answers, using temperature zero, disabled thinking, and a 1,024-token output limit. The JSON object contains \texttt{requirements}, with one to three entries. Each entry contains nonempty \texttt{answer}, \texttt{scope}, and \texttt{retrieval\_query} fields. The first describes the requested unknown information, the second its visual scope, and the last the neutral search phrase used for matching. Object modifiers and spatial conditions are preserved without guessing answer values or adding unasked attributes. Figure~3 illustrates possible types of visual evidence rather than literal parser outputs; in implementation, attributes are included only when requested by the question, without inferring unknown target attributes. Requirement count follows question content rather than a fixed count for identity, list, or counting questions.

Validation accepts JSON with or without a code fence, limits each field to 800 characters, and rejects duplicate search phrases after trimming whitespace and ignoring case. Invalid outputs use the complete question as one requirement. This fallback is distinct from correctly extracting one requirement.

\noindent\textbf{Quality aggregation.} Observation quality is $Q(o,f)=\prod_{k=1}^{6}\tilde x_k(o,f)^{w_k}$, with threshold $\theta_t(o)=\rho\max_{f\in\mathcal H_t(o)}Q(o,f)$. RGB images are resized to a longest side of 512 pixels using bilinear interpolation; masks use nearest-neighbor interpolation. Write $c(z)=\min(1,\max(0,z))$.

\begin{table}[t]\centering
\caption{Quality cues, weights, and positive floors.}\label{tab:app-cues-definition}
\small\setlength{\tabcolsep}{4pt}\renewcommand{\arraystretch}{1.12}
\begin{tabularx}{\linewidth}{@{}>{\hsize=0.88889\hsize\linewidth=\hsize\raggedright\arraybackslash}X>{\hsize=0.57143\hsize\linewidth=\hsize\raggedright\arraybackslash}X>{\hsize=0.57143\hsize\linewidth=\hsize\raggedright\arraybackslash}X>{\hsize=1.96825\hsize\linewidth=\hsize\raggedright\arraybackslash}X@{}}
\toprule
\textbf{Cue} & \textbf{Weight} & \textbf{Floor} & \textbf{Definition} \\
\midrule
Sharpness & 0.25 & 0.25 & Log Laplacian-to-image variance ratio, calibrated and clipped \\
Object area & 0.25 & 0.05 & Saturating square-root mask area \\
Boundary quality & 0.125 & 0.5 & Penalty near an image boundary \\
Detection confidence & 0.125 & $10^{-6}$ & Detector confidence with a positive floor \\
Depth visibility & 0.125 & 0.5 & Penalty for nearer non-object pixels inside the box \\
Text readability & 0.125 & 0.8 & Candidate height, contrast, and sharpness \\
\bottomrule
\end{tabularx}
\end{table}

Sharpness uses grayscale values in $[0,1]$, a $3\times3$ Gaussian filter with standard deviation 0.8, and a $3\times3$ Laplacian. The mask is eroded by a $3\times3$ kernel, reverting to the original mask if fewer than 16 pixels remain. The statistic and normalized cue are

\[
\begin{gathered}b=\log\!\left(1+\frac{\operatorname{Var}(\nabla^2 I)}{\operatorname{Var}(I)+10^{-4}}\right),\\\tilde x_1=0.25+0.75c(b/b_{\mathrm{ref}}).\end{gathered}
\]

The calibration takes the first six histories in calibration metadata order and the first four available grounding records per history in filename order. It pools valid object measurements, requires at least 20 measurements, and fixes their 90th percentile as the reference. Questions and answers are not used. Unavailable sharpness takes the floor 0.25.

For resized mask area $a$ and minimum box-to-image-boundary distance $m$,

\[
\begin{aligned}\tilde x_2&=0.05+0.95\min(1,\sqrt{a/1024}),\\\tilde x_3&=1-0.5c(1-m/4),\\\tilde x_4&=\max(10^{-6},p).\end{aligned}
\]

Depth is valid between 0.1 and 10 meters. Visibility requires at least 16 valid pixels and 25\% valid depth in both mask and box. With median object depth $z$, $\delta=\max(0.05,2\operatorname{MAD})$, and fraction $v$ of valid box pixels outside the mask whose depth is below $z-\delta$, use $\tilde x_5=1-0.5v$; unavailable visibility takes 0.5.

MSER candidates have heights of 4--64 pixels, width-to-height ratios of 0.1--5, fill ratios of 0.1--0.9, and at least half of their pixels inside the object mask. MSER uses delta 5, area 5--4096, maximum variation 0.25, and minimum diversity 0.2. Candidate height, grayscale 90th-minus-10th percentile contrast, and sharpness are denoted by $h_j,k_j,b_j$:

\[
\tilde x_6=0.8+0.2\max_j\left[c\!\left(\frac{h_j-4}{12}\right)c\!\left(\frac{k_j}{0.5}\right)c\!\left(\frac{b_j}{b_{\mathrm{ref}}}\right)\right]^{1/3}.
\]

The maximum is zero without a valid candidate. This cue estimates readability rather than recognizing text semantics.

\noindent\textbf{Scene groups.} A group uses its creating frame's unit global feature as a fixed representative. An incoming frame joins every group with cosine similarity at least $\eta$, creating a group only if none matches. Memberships are fixed and may overlap. Coverage retains at least one associated frame per group; the creating frame need not remain. There are no centroid updates or periodic regrouping.

\noindent\textbf{Construction and local maintenance.} Frames are processed in recorded order, one incoming frame per update. Greedy selection first maximizes newly covered constraints, then auxiliary retention utility, then source-frame ID. The utility averages best retained-to-historical ratios over recorded object and scene entries; scene entries use unit quality. This bookkeeping differs from the object-only evaluation in Section~\ref{app:protocol}.

The local pool includes incoming frames and memory frames sharing raw object or scene associations with them, before quality filtering. Other memory frames remain fixed. Local selection covers every constraint unmet by the fixed set. Pruning considers locally selected frames only, preferring smaller utility loss and then source-frame ID, and preserves complete coverage at every removal. The final memory is checked against all updated constraints.
For the maintenance comparison, global reselection operates
on the full update pool
$\mathcal{M}_t \cup \mathcal{B}_{t+1}$,
whereas local maintenance reselects only affected frames
while keeping the remaining memory fixed.
This full-pool baseline differs from reconstruction
over the entire observation history $\mathcal{H}_{t+1}$.
Equation~(3) defines the full-history construction
objective, whereas streaming maintenance operates
on $\mathcal{M}_t \cup \mathcal{B}_{t+1}$.
The incremental procedure guarantees coverage
feasibility under updated constraints, rather than
global memory-size optimality over the full history.

\noindent\textbf{Matching and normalization.} Global, crop, and description similarities are independently min--max normalized over the initial candidate pool of each selection run; ranges at most $10^{-8}$ map to $0.5$. These normalized scores remain fixed throughout greedy selection. With regions $\mathcal B(f)$,

\[
e_x(f)=\tfrac12\widehat g_x(f)+\tfrac12\max_{b\in\mathcal B(f)}\frac{\widehat c_x(b)+\widehat\ell_x(b)}{2}.
\]

Without regions, $e_x(f)=\widehat g_x(f)$. Relevance uses the complete question, $r_q(f)=\widehat g_q(f)$; requirement support is $s_q(f,d)=e_d(f)$. Both are fixed within a selection run. Current support and additional support are

\[
u_{q,d}(S)=\max_{g\in S}s_q(g,d),
\]

We initialize $u_{q,d}(\varnothing)=0$.

\[
\Delta_q(f\mid S)=\frac{1}{|\mathcal D(q)|}\sum_{d\in\mathcal D(q)}[s_q(f,d)-u_{q,d}(S)]_+.
\]

\[
\Phi_q(f\mid S)=\alpha r_q(f)+(1-\alpha)\Delta_q(f\mid S)-\lambda d_q(f,S).
\]

Pairwise cosine similarities of global features are clipped to $[0,1]$, shifted by the minimum over distinct candidate pairs, and divided by their range with a $10^{-8}$ floor. Redundancy is the maximum normalized similarity to a selected frame and is zero for an empty set. Normalization statistics are computed at the start of each selection run using its initial candidate pool and the corresponding similarity scores. They remain fixed throughout greedy selection and are not recomputed after each frame is selected. The subscript $q$ identifies the selection instance; the pairwise redundancy computation itself does not use question text.

\begin{table}[t]\centering
\caption{Default coverage and selection parameters.}\label{tab:params}
\small\setlength{\tabcolsep}{4pt}\renewcommand{\arraystretch}{1.12}
\begin{tabularx}{\linewidth}{@{}>{\hsize=1.00000\hsize\linewidth=\hsize\raggedright\arraybackslash}X>{\hsize=1.00000\hsize\linewidth=\hsize\raggedright\arraybackslash}X>{\hsize=1.00000\hsize\linewidth=\hsize\raggedright\arraybackslash}X>{\hsize=1.00000\hsize\linewidth=\hsize\raggedright\arraybackslash}X>{\hsize=1.00000\hsize\linewidth=\hsize\raggedright\arraybackslash}X>{\hsize=1.00000\hsize\linewidth=\hsize\raggedright\arraybackslash}X@{}}
\toprule
\textbf{$\rho$} & \textbf{$\eta$} & \textbf{$\alpha$} & \textbf{$\lambda$} & \textbf{$\tau$} & \textbf{$K$} \\
\midrule
0.70 & 0.80 & 0.50 & 0.10 & 0.05 & 3 \\
\bottomrule
\end{tabularx}
\end{table}

Scores are rounded to 12 decimals before source-frame-ID tie-breaking. Selection stops when the best score is strictly below $\tau$, when the budget is reached, or when candidates are exhausted. Equality to $\tau$ does not stop selection. Only current support and redundancy change across steps; answering-model reasoning does not feed back into selection.

Static-support controls replace $\Delta_q(f\mid S)$ by $\Delta_q(f\mid\varnothing)$ while still updating redundancy, and the fixed-three-frame control disables only score-based stopping.

\subsection{Fair comparison with existing methods}\label{app:baselines}

We ensure fair reproduction through fixed evaluation subsets, a shared answering backbone and evaluator, and preservation of each method's defining representation and retrieval mechanism.
All compared baselines use their best verified configurations under the common evaluation protocol.

Methods with an explicit frame-budget interface use at most three source frames. Other methods retain their input procedures and report actual source-frame counts. Caption methods supply scene text; this does not imply zero visual preprocessing. VLM-Grounder uses collages whose constituent source frames determine its visual input count. The main table's $100\%^{*}$ entries denote complete candidate-pool availability. Uniform sampling takes distinct frames from the original history without CoverMem filtering or relevance ranking; Blind receives question text only.

\subsection{Computational cost, timing, and storage scope}\label{app:cost}

The construction-stage breakdown is reported in main-paper Table~\ref*{tab:table6}.

Construction stages use NVIDIA L20 GPUs, with one GPU per computational stage; parsing, answering, and evaluation use remote APIs. Construction time is cumulative across the listed stages. Maintenance uses cached inputs and excludes raw perception and API waiting, with two runs per history and maintenance rule over 51 histories. These measurements distinguish reusable preprocessing, incremental maintenance, and question-time operations. Although CoverMem reduces answer-input tokens relative to uniform sampling, its additional parsing calls make total question-time API-token usage slightly higher; the token advantage is instead observed relative to ToolMerge.

The retained-frame ratio measures active source-frame counts rather than total storage: features, descriptions, associations, and preprocessing artifacts also occupy space. Cached maintenance time is not end-to-end query latency, and cumulative construction time is not simultaneous per-GPU utilization. History, frame, and region IDs bind caches to observations; file checksums and fixed configurations support reuse. Reference answers are excluded from construction and selection.

\FloatBarrier
\section{Additional experiments and analyses}\label{app:B}
\setcounter{table}{0}\renewcommand{\thetable}{B\arabic{table}}
\setcounter{figure}{0}\renewcommand{\thefigure}{B\arabic{figure}}

\subsection{Parameter sensitivity and quality design}\label{app:quality-sensitivity}

\begin{figure*}[!tbp]\centering
\includegraphics[width=0.92\textwidth]{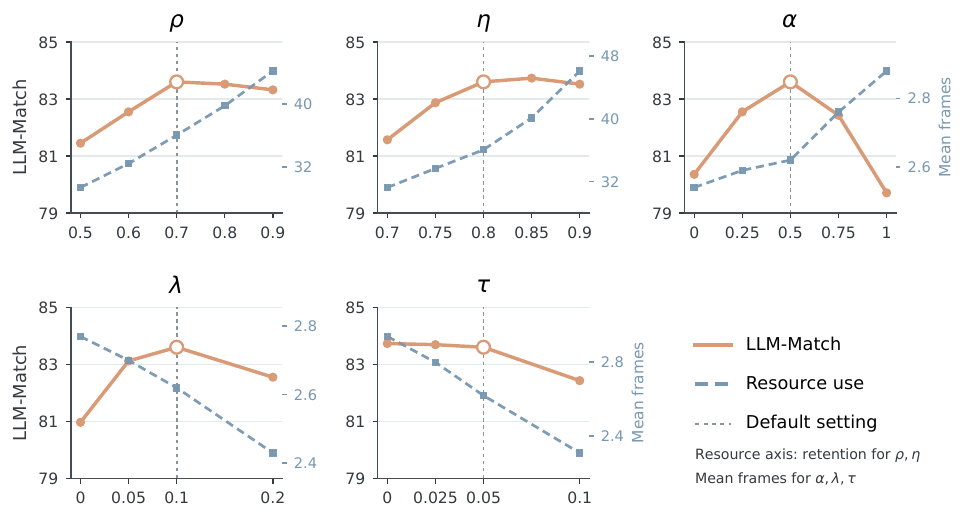}
\caption{One-factor sensitivity of five parameters. Orange denotes LLM-Match, and blue denotes resource use (retained frames for $\rho,\eta$; input frames for $\alpha,\lambda,\tau$). Dashed vertical lines mark default settings.}\label{fig:app-sensitivity}
\end{figure*}

Unless stated otherwise, supplementary analyses use the 616 OpenEQA ScanNet questions, main answering and evaluation models, and $K=3$. One parameter varies at a time (Figure~\ref{fig:app-sensitivity}). Changing $\rho$ or $\eta$ rebuilds memory, whereas changing selection weights or stopping reuses the default memory.
Default hyperparameters were fixed based on methodological
considerations before the main evaluation, without using
evaluation-set QA performance for parameter selection.
The same default configuration is applied across all
three benchmarks without benchmark-specific tuning.
The one-factor sensitivity analyses are conducted
post hoc to examine robustness and
performance--efficiency trade-offs, rather than
to optimize the default configuration.

\begin{table}[t]\centering
\caption{Construction-parameter sensitivity.}\label{tab:app-build-sens}
\small\setlength{\tabcolsep}{4pt}\renewcommand{\arraystretch}{1.12}
\begin{tabularx}{\linewidth}{@{}>{\hsize=1.00000\hsize\linewidth=\hsize\raggedright\arraybackslash}X>{\hsize=1.00000\hsize\linewidth=\hsize\raggedright\arraybackslash}X>{\hsize=1.00000\hsize\linewidth=\hsize\raggedright\arraybackslash}X>{\hsize=1.00000\hsize\linewidth=\hsize\raggedright\arraybackslash}X@{}}
\toprule
\textbf{Parameter} & \textbf{Value} & \textbf{LLM-Match} & \textbf{Retained (\%)} \\
\midrule
$\rho$ & 0.50 & 81.45 & 29.43 \\
$\rho$ & 0.60 & 82.55 & 32.46 \\
$\rho$ & 0.70 & 83.60 & 36.05 \\
$\rho$ & 0.80 & 83.52 & 39.77 \\
$\rho$ & 0.90 & 83.32 & 44.16 \\
$\eta$ & 0.70 & 81.57 & 31.23 \\
$\eta$ & 0.75 & 82.87 & 33.66 \\
$\eta$ & 0.80 & 83.60 & 36.05 \\
$\eta$ & 0.85 & 83.73 & 40.11 \\
$\eta$ & 0.90 & 83.52 & 46.09 \\
\bottomrule
\end{tabularx}
\end{table}

Increasing $\rho$ from 0.5 to 0.7 raises answer quality alongside retention; further increases retain more frames without a corresponding QA improvement. Setting $\eta=0.85$ slightly exceeds the default score but uses more memory. The defaults balance quality and compactness rather than maximizing every isolated score.

\begin{table}[t]\centering
\caption{Selection-parameter sensitivity.}\label{tab:app-select-sens}
\small\setlength{\tabcolsep}{4pt}\renewcommand{\arraystretch}{1.12}
\begin{tabularx}{\linewidth}{@{}>{\hsize=1.00000\hsize\linewidth=\hsize\raggedright\arraybackslash}X>{\hsize=1.00000\hsize\linewidth=\hsize\raggedright\arraybackslash}X>{\hsize=1.00000\hsize\linewidth=\hsize\raggedright\arraybackslash}X>{\hsize=1.00000\hsize\linewidth=\hsize\raggedright\arraybackslash}X@{}}
\toprule
\textbf{Parameter} & \textbf{Value} & \textbf{LLM-Match} & \textbf{Input frames} \\
\midrule
$\alpha$ & 0 & 80.36 & 2.54 \\
$\alpha$ & 0.25 & 82.55 & 2.59 \\
$\alpha$ & 0.50 & 83.60 & 2.62 \\
$\alpha$ & 0.75 & 82.43 & 2.76 \\
$\alpha$ & 1 & 79.71 & 2.88 \\
$\lambda$ & 0 & 80.97 & 2.77 \\
$\lambda$ & 0.05 & 83.12 & 2.70 \\
$\lambda$ & 0.10 & 83.60 & 2.62 \\
$\lambda$ & 0.20 & 82.55 & 2.43 \\
$\tau$ & 0 & 83.73 & 2.94 \\
$\tau$ & 0.025 & 83.69 & 2.80 \\
$\tau$ & 0.05 & 83.60 & 2.62 \\
$\tau$ & 0.10 & 82.43 & 2.31 \\
\bottomrule
\end{tabularx}
\end{table}

Moderate $\alpha$ balances question relevance and new support.
Setting $\alpha=1$ both changes the relevance scaling and disables requirement support; therefore, it is not equivalent to every separately defined support ablation. Setting $\lambda=0$ reproduces the main-paper no-redundancy score of 80.97. Excessive redundancy penalties can suppress similar views that still reveal useful details.

Lowering $\tau$ increases actual input size: $\tau=0$ reaches 83.73 using 2.94 frames on average, compared with 83.60 and 2.62 at the default. Section~\ref{app:parser-evidence} separately reports a fixed-three-frame control. Nearby parameter settings have comparable performance, while stricter memory constraints increase resource use.

\noindent\textbf{Effect of source-frame budget.}
The non-monotonic trends in main-paper Figure~7
reflect that additional visual inputs do not
necessarily improve answer quality.
For uniform sampling, increasing the budget may
introduce redundant or distracting observations,
potentially affecting VLM responses.
ToolMerge exhibits a sharp improvement from
$K=2$ to $K=3$, consistent with the possibility
that an additional frame supplies complementary
evidence for multi-part questions.
These observations suggest a nonlinear relationship
between visual input budgets and answer quality,
rather than guaranteed monotonic improvements.

\noindent\textbf{Individual cue removal.} For each cue $k$, set $w_k=0$ and renormalize the remaining weights by $w_j/(1-w_k)$. Recompute qualities, historical thresholds, and memory while keeping downstream selection and budget fixed. Qualified-object ratio and relative quality always use the original full six-cue score with the common 0.7 evaluation threshold.

\begin{table*}[t]\centering
\caption{Leave-one-cue-out construction ablations under a common full-quality evaluation.}\label{tab:app-cue-removal}
\small\setlength{\tabcolsep}{4pt}\renewcommand{\arraystretch}{1.12}
\begin{tabularx}{\linewidth}{@{}>{\hsize=1.62857\hsize\linewidth=\hsize\raggedright\arraybackslash}X>{\hsize=0.85714\hsize\linewidth=\hsize\raggedright\arraybackslash}X>{\hsize=0.85714\hsize\linewidth=\hsize\raggedright\arraybackslash}X>{\hsize=0.85714\hsize\linewidth=\hsize\raggedright\arraybackslash}X>{\hsize=0.94286\hsize\linewidth=\hsize\raggedright\arraybackslash}X>{\hsize=0.85714\hsize\linewidth=\hsize\raggedright\arraybackslash}X@{}}
\toprule
\textbf{Configuration} & \textbf{LLM-Match} & \textbf{Retained (\%)} & \textbf{Qualified (\%)} & \textbf{Rel. quality (\%)} & \textbf{Input frames} \\
\midrule
Full six cues & 83.60 & 36.05 & 100 & 98.49 & 2.62 \\
No sharpness & 81.74 & 33.55 & 96.84 & 96.72 & 2.65 \\
No area & 81.29 & 34.15 & 96.08 & 96.11 & 2.67 \\
No boundary quality & 82.67 & 35.12 & 98.23 & 97.65 & 2.63 \\
No confidence & 83.00 & 35.48 & 98.56 & 97.91 & 2.62 \\
No visibility & 82.39 & 34.75 & 97.41 & 97.32 & 2.64 \\
No readability & 83.64 & 35.92 & 99.67 & 98.37 & 2.62 \\
\bottomrule
\end{tabularx}
\end{table*}

\begin{figure*}[!tbp]\centering
\includegraphics[width=0.90\textwidth]{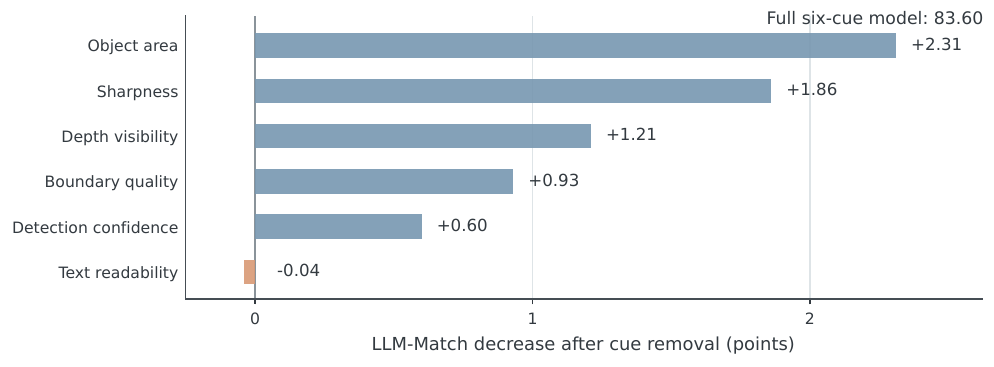}
\caption{LLM-Match decrease after removing each quality cue, relative to the full six-cue configuration (83.60). Negative values indicate an improvement.}\label{fig:app-cues}
\end{figure*}

As shown in Figure~\ref{fig:app-cues}, removing area or sharpness causes the largest decreases, consistent with object scale and clarity being prominent quality factors here. Boundary and visibility cues help distinguish truncated and occluded observations, while confidence adds a smaller gain. Removing readability changes QA from 83.60 to 83.64, about one rating-unit step on this subset. The small change indicates a limited marginal QA contribution on this subset.

The four configurations in main-paper Figure~5
compare different quality-scoring strategies,
among which the default MCQ achieves the highest score.
The leave-one-cue-out experiments in Table~B3
examine individual cue contributions separately;
removing readability yields a negligible
0.04-point difference, indicating limited
sensitivity to this cue on the evaluated subset.

\subsection{Statistical significance}\label{app:significance}

We retain the archived main-paper comparisons in Table~\ref{tab:significance}. Differences are CoverMem minus the comparator. We use 10,000 paired scene-bootstrap resamples for 95\% percentile confidence intervals and 100,000 two-sided paired permutations that swap method labels jointly within each physical scene. Holm correction is applied across the four comparisons. Questions from different histories of the same scene remain grouped.

\begin{table}[t]\centering
\caption{Archived significance results for the main-paper contrasts. Confidence intervals are unadjusted; p-values are Holm-adjusted across four comparisons.}\label{tab:significance}
\small\setlength{\tabcolsep}{4pt}\renewcommand{\arraystretch}{1.12}
\begin{tabularx}{\linewidth}{@{}>{\hsize=1.19403\hsize\linewidth=\hsize\raggedright\arraybackslash}X>{\hsize=1.19403\hsize\linewidth=\hsize\raggedright\arraybackslash}X>{\hsize=0.52239\hsize\linewidth=\hsize\raggedright\arraybackslash}X>{\hsize=1.34328\hsize\linewidth=\hsize\raggedright\arraybackslash}X>{\hsize=0.74627\hsize\linewidth=\hsize\raggedright\arraybackslash}X@{}}
\toprule
\textbf{Benchmark} & \textbf{Reference} & \textbf{$\Delta$} & \textbf{95\% CI} & \textbf{Holm $p$} \\
\midrule
OpenEQA ScanNet & ToolMerge & +4.14 & [+1.50, +6.80] & 0.006 \\
EXPRESS & ToolMerge & +1.03 & [+0.15, +1.90] & 0.030 \\
ScanQA & ToolMerge & +6.66 & [+3.80, +9.50] & 0.0015 \\
OpenEQA ScanNet & Size-matched Uniform memory & +7.50 & [+4.80, +10.20] & 0.0004 \\
\bottomrule
\end{tabularx}
\end{table}

All four archived intervals are above zero and adjusted p-values are at most 0.03, matching the main-paper statement. EXPRESS has a smaller effect than the other benchmarks. The uncertainty analysis conditions on the existing answer--judge outputs and sampled scenes; it does not estimate variability from repeated parsing, answering, or evaluation API calls. Statistical inference remains tied to these specific comparisons and resampling units.

\subsection{Parser reliability and evidence adequacy}\label{app:parser-evidence}

\noindent\textbf{Format and semantic consistency.} Format checks use the parser schema. A separate semantic consistency checker using GPT-6 Luna sees only the original question and extracted requirements, not selected images, method scores, or downstream answers. It checks preservation of requested information, modifiers, and relations, and flags unasked content or guessed answer values.

\begin{table}[t]\centering
\caption{Parsing diagnostics on all 616 questions. Requirement counts include whole-question fallbacks.}\label{tab:app-parser}
\small\setlength{\tabcolsep}{4pt}\renewcommand{\arraystretch}{1.12}
\begin{tabularx}{\linewidth}{@{}>{\hsize=1.80000\hsize\linewidth=\hsize\raggedright\arraybackslash}X>{\hsize=0.60000\hsize\linewidth=\hsize\raggedright\arraybackslash}X>{\hsize=0.60000\hsize\linewidth=\hsize\raggedright\arraybackslash}X@{}}
\toprule
\textbf{Diagnostic} & \textbf{Count} & \textbf{Percent} \\
\midrule
Schema-valid outputs & 613 & 99.51 \\
Fallbacks & 3 & 0.49 \\
Final one requirement & 216 & 35.06 \\
Final two requirements & 271 & 43.99 \\
Final three requirements & 129 & 20.94 \\
Automatic semantic check passed & 580 & 94.16 \\
\bottomrule
\end{tabularx}
\end{table}

\begin{table}[t]\centering
\caption{Primary automatically flagged semantic issue, assigning one category per flagged question.}\label{tab:app-parser-errors}
\small\setlength{\tabcolsep}{4pt}\renewcommand{\arraystretch}{1.12}
\begin{tabularx}{\linewidth}{@{}>{\hsize=1.60000\hsize\linewidth=\hsize\raggedright\arraybackslash}X>{\hsize=0.40000\hsize\linewidth=\hsize\raggedright\arraybackslash}X@{}}
\toprule
\textbf{Primary issue} & \textbf{Count / 36} \\
\midrule
Omitted requested information & 14 \\
Weakened relation or modifier & 11 \\
Added unasked information & 7 \\
Guessed answer value & 4 \\
\bottomrule
\end{tabularx}
\end{table}

The 580 passed questions score 85.86, whereas the 36 flagged questions score 47.22, aggregating to 83.60 over the full subset. Since question difficulty may differ between groups, this split describes association rather than a causal effect of parsing errors.

\begin{table}[t]\centering
\caption{Separate parser and description interventions with memory and visual features fixed.}\label{tab:app-repair}
\small\setlength{\tabcolsep}{4pt}\renewcommand{\arraystretch}{1.12}
\begin{tabularx}{\linewidth}{@{}>{\hsize=1.60000\hsize\linewidth=\hsize\raggedright\arraybackslash}X>{\hsize=0.40000\hsize\linewidth=\hsize\raggedright\arraybackslash}X@{}}
\toprule
\textbf{Intervention} & \textbf{LLM-Match} \\
\midrule
Original pipeline & 83.60 \\
Correct only automatically identified parser issues & 84.21 \\
Correct only automatically identified description issues & 83.97 \\
\bottomrule
\end{tabularx}
\end{table}

Parser corrections restore information already specified by the question without supplying unknown answer values. Description corrections use region appearance. These interventions are separate and their gains are not added. Correcting parser issues improves QA by approximately 0.61 points; recognition and relation interpretation can remain difficult afterward.

\noindent\textbf{Independent evidence assessment.} Information needs are extracted once from the original question alone, before any images are evaluated, and fixed across all methods and pipeline stages. GPT-6 Luna then independently assesses each image set without access to method names, selector scores, or answering-model outputs. Counting, lists, and relations may require joint evidence. For knowledge questions, the visual assessment concerns their visual premises rather than requiring an image to supply external facts.

\[
\mathrm{ReqCov}(q,S)=\frac{1}{|\mathcal D^*(q)|}\sum_{d\in\mathcal D^*(q)}a_{q,d}(S).
\]

Here $a_{q,d}(S)=1$ when the selected frames provide visible evidence for need $d$, and zero otherwise.

The independently derived $\mathcal D^*(q)$ avoids shrinking the evaluation denominator when the selector omits a difficult need. Set-level adequacy $A(q,S)$ examines the selected set jointly, allowing multiple frames to support one need. These diagnostics follow the automatic-evaluation scope in Section~\ref{app:protocol}.

\begin{table}[t]\centering
\caption{Selection and independent evidence diagnostics on the same default memory.}\label{tab:app-adequacy}
\small\setlength{\tabcolsep}{4pt}\renewcommand{\arraystretch}{1.12}
\begin{tabularx}{\linewidth}{@{}>{\hsize=1.74242\hsize\linewidth=\hsize\raggedright\arraybackslash}X>{\hsize=0.75758\hsize\linewidth=\hsize\raggedright\arraybackslash}X>{\hsize=0.83333\hsize\linewidth=\hsize\raggedright\arraybackslash}X>{\hsize=0.90909\hsize\linewidth=\hsize\raggedright\arraybackslash}X>{\hsize=0.75758\hsize\linewidth=\hsize\raggedright\arraybackslash}X@{}}
\toprule
\textbf{Selection} & \textbf{LLM-Match} & \textbf{ReqCov (\%)} & \textbf{Set adequate (\%)} & \textbf{Input frames} \\
\midrule
Relevance only & 76.54 & 65.10 & 59.74 & 2.73 \\
Static requirement support & 81.86 & 72.08 & 67.69 & 2.69 \\
Dynamic residual support & 83.60 & 76.30 & 72.56 & 2.62 \\
Dynamic support, exactly three frames & 83.73 & 77.11 & 73.05 & 3.00 \\
\bottomrule
\end{tabularx}
\end{table}

Static and dynamic support share requirements, candidates, and matching scores. Dynamic updates improve both coverage and adequacy while using fewer input frames. Exactly three frames improve coverage further but add little QA relative to the default, consistent with the main-paper diminishing returns from additional frames.

\begin{figure*}[!tbp]\centering
\includegraphics[width=0.93\textwidth]{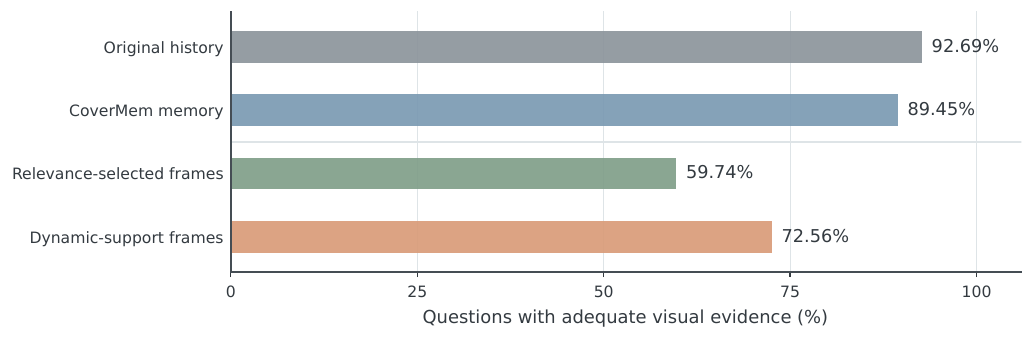}
\caption{Set-level evidence adequacy across memory and selection stages (616 questions). Adequate-question counts are 571, 551, 368, and 447 in plotted order; the last two groups compare selection strategies on the same memory.}\label{fig:app-stages}
\end{figure*}

The independent assessment (Figure~\ref{fig:app-stages}) identifies adequate evidence
for 571 of 616 questions (92.69\%) in the original
histories and 551 questions (89.45\%) in CoverMem
memory, corresponding to a decrease of only
3.25 percentage points in aggregate evidence adequacy.
After budgeted selection, 447 questions (72.56\%)
are assessed as having adequate visual evidence, suggesting that
question-time evidence allocation remains an important
bottleneck. Evidence absent from the original history
cannot be recovered by selection.

\begin{table}[t]\centering
\caption{Question-type evidence adequacy alongside the main-paper QA scores.}\label{tab:app-type-adequacy}
\small\setlength{\tabcolsep}{4pt}\renewcommand{\arraystretch}{1.12}
\begin{tabularx}{\linewidth}{@{}>{\hsize=1.62963\hsize\linewidth=\hsize\raggedright\arraybackslash}X>{\hsize=0.74074\hsize\linewidth=\hsize\raggedright\arraybackslash}X>{\hsize=0.74074\hsize\linewidth=\hsize\raggedright\arraybackslash}X>{\hsize=0.88889\hsize\linewidth=\hsize\raggedright\arraybackslash}X@{}}
\toprule
\textbf{Question type} & \textbf{Questions} & \textbf{Main QA} & \textbf{Set adequate (\%)} \\
\midrule
Attribute recognition & 90 & 85.28 & 73.33 \\
Function reasoning & 85 & 88.53 & 80.00 \\
Object localization & 92 & 88.32 & 76.09 \\
Object recognition & 90 & 79.44 & 66.67 \\
State recognition & 94 & 83.24 & 76.60 \\
Spatial understanding & 85 & 71.18 & 52.94 \\
World knowledge & 80 & 89.38 & 82.50 \\
\bottomrule
\end{tabularx}
\end{table}

The seven categories (Table~\ref{tab:app-type-adequacy}) contain 616 questions and 447 adequate selected sets. Spatial understanding has comparatively low evidence adequacy and main-paper QA. Missing relation evidence and answering-model interpretation errors are distinct contributors, rather than all spatial errors being attributable to visual redundancy.

\begin{figure*}[!tbp]\centering
\includegraphics[width=0.88\textwidth]{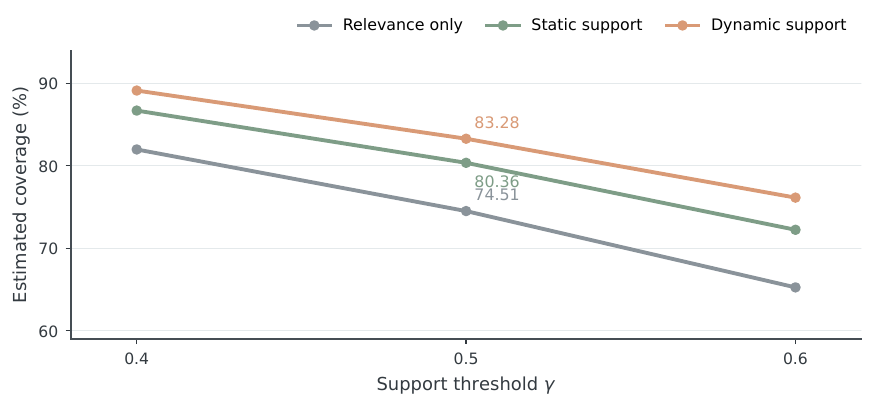}
\caption{Estimated requirement coverage as the support threshold $\gamma$ increases. This selector-score diagnostic complements independent visual evidence assessment.}\label{fig:app-support-threshold}
\end{figure*}

This auxiliary measure (Figure~\ref{fig:app-support-threshold}) declares support when $\max_{f\in S}s_q(f,d)\geq\gamma$. Dynamic support is higher at all three thresholds, but the metric uses the selector's own scores and is distinct from Table~\ref{tab:app-adequacy}. Normalized matching scores are not calibrated adequacy probabilities.

\subsection{Associations, maintenance, and stream growth}\label{app:maintenance-diagnostics}

As reported in main-paper Table~3, local maintenance
and full-pool reselection produce identical final memory
frame sets, question answers, and scores on the
51 unperturbed histories.
Both methods operate on the same update pool
$\mathcal{M}_t \cup \mathcal{B}_{t+1}$ rather
than the complete history $\mathcal{H}_{t+1}$.
Local maintenance reconsiders only 49.30\%
of these candidates, reducing cached maintenance
time from 3.763 to 1.958 seconds per history.
Thus, the reported 48.0\% reduction is relative
to full-update-pool reselection, not complete
history reconstruction or end-to-end processing.

Reselection uses existing memory plus incoming frames as its denominator, whereas retention uses the original history. The time decrease is 47.97\%, rounded to the main-paper 48.0\%. Timing and storage definitions are consolidated in Section~\ref{app:cost}.

\noindent\textbf{Association perturbations.} Merges, splits, and missing observation associations are introduced into cached streams before replay. Both maintenance rules consume the same perturbed stream; old associations are not retroactively changed during replay. Perturbation percentages refer to the selected identifiers or observation associations, with each intervention performed separately.

\begin{table}[t]\centering
\caption{\textbf{Association perturbations.}
L/G set match measures exact final-memory agreement
between local maintenance and full-pool reselection
across 51 histories under the same perturbation.}
\label{tab:app-association}
\small\setlength{\tabcolsep}{4pt}\renewcommand{\arraystretch}{1.12}
\begin{tabularx}{\linewidth}{@{}>{\hsize=1.61538\hsize\linewidth=\hsize\raggedright\arraybackslash}X>{\hsize=0.76923\hsize\linewidth=\hsize\raggedright\arraybackslash}X>{\hsize=0.76923\hsize\linewidth=\hsize\raggedright\arraybackslash}X>{\hsize=0.92308\hsize\linewidth=\hsize\raggedright\arraybackslash}X>{\hsize=0.92308\hsize\linewidth=\hsize\raggedright\arraybackslash}X@{}}
\toprule
\textbf{Perturbation} & \textbf{LLM-Match} & \textbf{Ret. (\%)} & \textbf{Ref. qualified (\%)} & \textbf{
L/G set match (\%)
} \\
\midrule
None & 83.60 & 36.05 & 100 & 100 \\
Identifier merge, 5\% & 82.43 & 34.88 & 97.68 & 96.08 \\
Identifier merge, 10\% & 80.97 & 33.26 & 94.17 & 90.20 \\
Identifier split, 5\% & 83.20 & 38.44 & 100 & 98.04 \\
Identifier split, 10\% & 82.87 & 40.72 & 100 & 94.12 \\
Association dropout, 5\% & 82.18 & 36.78 & 97.36 & 96.08 \\
Association dropout, 10\% & 80.68 & 37.62 & 94.82 & 90.20 \\
\bottomrule
\end{tabularx}
\end{table}

Each run satisfies the coverage constraints induced by its perturbed associations. LLM-Match, retention, and reference qualification
are reported for local maintenance, with reference qualification evaluated against the original unperturbed associations and full six-cue quality score. Merging may satisfy nominal coverage while losing distinctions between original targets; splitting increases target count and retention. Dropouts can remove useful witnesses. Agreement percentages of 100, 98.04, 96.08, 94.12, and 90.20 correspond to 51, 50, 49, 48, and 46 histories. Divergence under an intervention differs from the unperturbed main comparison and does not imply lost coverage.

\begin{figure*}[!tbp]\centering
\includegraphics[width=0.86\textwidth]{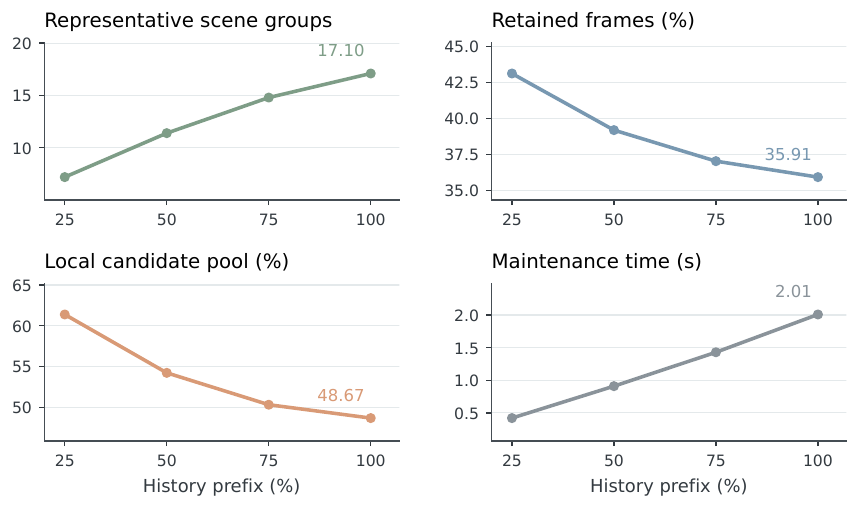}
\caption{Increasing prefixes of the ten longest available histories. Scene groups, retained-frame ratio, local candidate fraction, and maintenance time are averaged over the same histories.}\label{fig:app-stream}
\end{figure*}

Coverage is satisfied and local/global final-set agreement remains complete at these prefixes. As shown in Figure~\ref{fig:app-stream}, scene-group count grows with observed content while the local fraction decreases. This subset's final ratios differ from the full 51-history average because the aggregation sets differ.

\begin{samepage}
Residual coverage retains the submodular set-cover structure after fixing part of the memory~\cite{bach2013learning,chvatal1979greedy,wolsey1982analysis}; Section~\ref{app:coverage-properties} distinguishes coverage feasibility from equality of greedy outputs.\par
\end{samepage}

\subsection{Cross-benchmark component analyses}\label{app:cross-benchmark}

EXPRESS and ScanQA keep the main answering model, judge, and $K=3$. Plain object coverage replaces quality qualification by observed association without removing object constraints; removing scene coverage preserves full object-quality constraints. Each memory is rebuilt independently.

\begin{table}[t]\centering
\caption{EXPRESS construction ablations.}\label{tab:app-cross-express}
\small\setlength{\tabcolsep}{4pt}\renewcommand{\arraystretch}{1.12}
\begin{tabularx}{\linewidth}{@{}>{\hsize=1.66667\hsize\linewidth=\hsize\raggedright\arraybackslash}X>{\hsize=0.83333\hsize\linewidth=\hsize\raggedright\arraybackslash}X>{\hsize=0.83333\hsize\linewidth=\hsize\raggedright\arraybackslash}X>{\hsize=0.83333\hsize\linewidth=\hsize\raggedright\arraybackslash}X>{\hsize=0.83333\hsize\linewidth=\hsize\raggedright\arraybackslash}X@{}}
\toprule
\textbf{Configuration} & \textbf{LLM-Match} & \textbf{Ret. (\%)} & \textbf{Qual. (\%)} & \textbf{Input frames} \\
\midrule
Full method & 78.99 & 24.94 & 100 & 2.51 \\
Plain object coverage & 77.18 & 20.65 & 87.42 & 2.57 \\
Without scene coverage & 75.27 & 21.78 & 100 & 2.56 \\
\bottomrule
\end{tabularx}
\end{table}

\begin{table}[t]\centering
\caption{ScanQA construction ablations.}\label{tab:app-cross-scanqa}
\small\setlength{\tabcolsep}{4pt}\renewcommand{\arraystretch}{1.12}
\begin{tabularx}{\linewidth}{@{}>{\hsize=1.66667\hsize\linewidth=\hsize\raggedright\arraybackslash}X>{\hsize=0.83333\hsize\linewidth=\hsize\raggedright\arraybackslash}X>{\hsize=0.83333\hsize\linewidth=\hsize\raggedright\arraybackslash}X>{\hsize=0.83333\hsize\linewidth=\hsize\raggedright\arraybackslash}X>{\hsize=0.83333\hsize\linewidth=\hsize\raggedright\arraybackslash}X@{}}
\toprule
\textbf{Configuration} & \textbf{LLM-Match} & \textbf{Ret. (\%)} & \textbf{Qual. (\%)} & \textbf{Input frames} \\
\midrule
Full method & 79.68 & 22.40 & 100 & 3.00 \\
Plain object coverage & 77.30 & 18.72 & 88.76 & 3.00 \\
Without scene coverage & 75.79 & 19.53 & 100 & 3.00 \\
\bottomrule
\end{tabularx}
\end{table}

On both benchmarks, quality qualification trades additional retained frames for stronger quality witnesses. Removing scene coverage preserves qualified objects but lowers QA, indicating that object quality does not replace layout and context. As in the main construction ablation, coverage rules and resulting memory size change jointly. Therefore, these ablations do not isolate
a memory-size-controlled causal effect
of representative-scene coverage.

\begin{table}[t]\centering
\caption{Cross-benchmark selection ablations on the same full-method memory.}\label{tab:app-cross-selection}
\small\setlength{\tabcolsep}{4pt}\renewcommand{\arraystretch}{1.12}
\begin{tabularx}{\linewidth}{@{}>{\hsize=1.00000\hsize\linewidth=\hsize\raggedright\arraybackslash}X>{\hsize=1.00000\hsize\linewidth=\hsize\raggedright\arraybackslash}X>{\hsize=1.00000\hsize\linewidth=\hsize\raggedright\arraybackslash}X>{\hsize=1.00000\hsize\linewidth=\hsize\raggedright\arraybackslash}X@{}}
\toprule
\textbf{Benchmark} & \textbf{Relevance} & \textbf{Static support} & \textbf{Dynamic support} \\
\midrule
EXPRESS & 74.46 & 77.53 & 78.99 \\
ScanQA & 73.74 & 77.83 & 79.68 \\
\bottomrule
\end{tabularx}
\end{table}

Static and dynamic support use the same global and regional evidence, relevance, and redundancy. Their gains vary by benchmark; the dynamic margin is smaller on EXPRESS than ScanQA, consistent with the varying complete-pipeline improvements in the main paper.

\subsection{Controlled attribution of selection gains}\label{app:selection-attribution}

\begin{table}[t]\centering
\caption{Selection controls on the same memory with fixed relevance scores, redundancy penalty, and frame budget.}\label{tab:app-attribution}
\small\setlength{\tabcolsep}{4pt}\renewcommand{\arraystretch}{1.12}
\begin{tabularx}{\linewidth}{@{}>{\hsize=1.60000\hsize\linewidth=\hsize\raggedright\arraybackslash}X>{\hsize=0.40000\hsize\linewidth=\hsize\raggedright\arraybackslash}X@{}}
\toprule
\textbf{Requirements and support} & \textbf{LLM-Match} \\
\midrule
Relevance + redundancy & 79.95 \\
Whole question; static global support & 79.71 \\
Parsed requirements; static global support & 81.17 \\
Parsed requirements; static global + regional support & 81.86 \\
Parsed requirements; dynamic global + regional support & 83.60 \\
\bottomrule
\end{tabularx}
\end{table}

For the whole-question control, $\mathcal D(q)=\{q\}$ and static global support equals $r_q(f)$, so the score reduces to $r_q(f)-\lambda d_q(f,S)$. This is the same selector as \mbox{$\alpha=1$} in Section~\ref{app:quality-sensitivity}, and both use the reported result of 79.71. The relevance-plus-redundancy control instead keeps $0.5r_q(f)-\lambda d_q(f,S)$, giving a different relevance-to-redundancy weighting.

Replacing the whole question with parsed requirements in the static global-support control increases QA by 1.46 points. The main-paper increase of 1.22 uses the separate relevance-plus-redundancy baseline of 79.95. Static regional matching adds 0.69 and dynamic updating adds 1.74. The 7.06-point advantage over relevance-only selection concerns the combined selection configuration rather than decomposition alone.

Section~\ref{app:support-properties} relates additional support to a submodular objective and distinguishes its guarantees from the full selection rule~\cite{nemhauser1978analysis}.

\begin{table}[t]\centering
\caption{Regional branches within dynamic support.}\label{tab:app-regions}
\small\setlength{\tabcolsep}{4pt}\renewcommand{\arraystretch}{1.12}
\begin{tabularx}{\linewidth}{@{}>{\hsize=1.60000\hsize\linewidth=\hsize\raggedright\arraybackslash}X>{\hsize=0.40000\hsize\linewidth=\hsize\raggedright\arraybackslash}X@{}}
\toprule
\textbf{Support representation} & \textbf{LLM-Match} \\
\midrule
Global only & 82.26 \\
Global + crops & 83.04 \\
Global + descriptions & 82.79 \\
Global + crops + descriptions & 83.60 \\
\bottomrule
\end{tabularx}
\end{table}

Both regional branches improve on global-only support, and their combination scores highest among these branch controls. This complements the separate description intervention in Section~\ref{app:parser-evidence}.

\begin{table*}[t]\centering
\caption{Candidate-pool normalization control. Shared statistics are computed on the full-history reference pool.}\label{tab:app-normalization}
\small\setlength{\tabcolsep}{4pt}\renewcommand{\arraystretch}{1.12}
\begin{tabularx}{\linewidth}{@{}>{\hsize=1.00000\hsize\linewidth=\hsize\raggedright\arraybackslash}X>{\hsize=1.00000\hsize\linewidth=\hsize\raggedright\arraybackslash}X>{\hsize=1.00000\hsize\linewidth=\hsize\raggedright\arraybackslash}X>{\hsize=1.00000\hsize\linewidth=\hsize\raggedright\arraybackslash}X@{}}
\toprule
\textbf{Pool} & \textbf{Selection} & \textbf{Pool-wise QA} & \textbf{Shared-statistics QA} \\
\midrule
Full history & Relevance & 77.52 & 77.52 \\
Memory & Relevance & 76.54 & 76.62 \\
Full history & Complementary & 84.50 & 84.50 \\
Memory & Complementary & 83.60 & 83.52 \\
\bottomrule
\end{tabularx}
\end{table*}

Full-history outputs remain unchanged by the shared-statistics control. Memory scores shift slightly, while complementary selection remains stronger within each pool and full history remains above compressed memory. Thus, the principal ranking is retained when normalization is shared.

\section{Mathematical properties and scope}\label{app:C}
\setcounter{table}{0}\renewcommand{\thetable}{C\arabic{table}}
\setcounter{figure}{0}\renewcommand{\thefigure}{C\arabic{figure}}

\subsection{Coverage preservation and local maintenance}\label{app:coverage-properties}

With quality thresholds and associations fixed, define $\operatorname{Cov}_t(S)=\bigcup_{f\in S}\mathcal R_t(f)$ and $F_t(S)=|\operatorname{Cov}_t(S)|$. Object and scene constraints have distinct identifiers. For $A\subseteq T$,

\[
\begin{gathered}F_t(A\cup\{f\})-F_t(A)\\=|\mathcal R_t(f)\setminus\operatorname{Cov}_t(A)|\\\geq|\mathcal R_t(f)\setminus\operatorname{Cov}_t(T)|.\end{gathered}
\]

Hence coverage is normalized, monotone, and submodular, giving the classical greedy set-cover structure. Greedy construction need not produce the smallest possible memory.

\noindent\textbf{Proposition C1: feasible local updates.} Suppose old memory covers old constraints, old qualities and associations are immutable, $0<\rho\leq1$, historical maxima are updated from incoming frames, and the local pool and residual universe follow the main definition. Then the local pool contains a cover of all residual constraints; greedy selection and coverage-preserving local pruning yield feasible updated memory.

\noindent\textbf{Proof.} An unchanged historical maximum preserves its old qualified witness. An increased maximum is attained in the incoming batch and qualifies because $\rho\leq1$; new objects have the same kind of witness. Old scenes keep witnesses, and every new scene has a creating frame. Therefore $\mathcal M_t\cup\mathcal B_{t+1}$ covers the updated universe. Partition this union into fixed frames $P$ and locally selectable frames $W$. Every constraint uncovered by $P$ has a witness in $W$. While residual constraints remain, an unselected frame supplies positive coverage, so finite greedy selection terminates with a cover. Every pruning step preserves coverage together with $P$.

The residual universe includes old constraints supported by local frames, not only newly introduced constraints. Defining affected frames before quality filtering identifies an old witness even when an increased threshold disqualifies it.

\noindent\textbf{Local/global agreement.} Fixing part of the memory changes the feasible search and greedy order. Feasibility therefore does not imply equal membership or size. The main-paper equality is an observation on the evaluated unperturbed histories; the perturbation diagnostics examine a distinct regime.

\noindent\textbf{Scene-group growth.} A representative is created only if its similarity to every older representative is below $\eta$. Fixed features avoid centroid-update drift, but diverse observations can create further groups. The direct bound $|\mathcal G_t|\leq|\mathcal H_t|$ does not imply constant retained memory.

\subsection{Residual support and related selection objectives}\label{app:support-properties}

\[
U_q(S)=\frac{1}{|\mathcal D(q)|}\sum_{d\in\mathcal D(q)}\max_{f\in S}s_q(f,d),\qquad U_q(\varnothing)=0.
\]

For a fixed nonempty requirement set and fixed scores, $U_q$ is normalized, monotone, and submodular. Increasing the selected set can only increase its best support for each requirement and thus decrease a new frame's positive improvement. Its marginal is exactly

\[
\begin{gathered}U_q(S\cup\{f\})-U_q(S)\\=\frac{1}{|\mathcal D(q)|}\sum_d[s_q(f,d)-u_{q,d}(S)]_+\\=\Delta_q(f\mid S).\end{gathered}
\]

The per-demand maximum has facility-location form. CoverMem instantiates demand units with question requirements and couples their residual support with a memory retained before the question. The complete ranking additionally includes relevance, maximum visual similarity, and stopping; the support proposition alone does not give the deployed rule a monotonicity or $1-1/e$ approximation guarantee. Fixed pairwise quadratic penalties likewise differ from updating the best support of each demand. Here, complementarity refers to the marginal
improvement in the best single-frame support
for each question requirement. The selection
objective does not explicitly model evidence
that emerges only from jointly interpreting
multiple frames for the same requirement.
Such cross-frame reasoning may still be
performed by the downstream answering VLM.

\subsection{Positioning relative to structured memories}\label{app:structured-scope}

\begin{table}[t]\centering
\caption{Capabilities of full-frame and structured memories.}\label{tab:supp-memory-comparison}
\small\setlength{\tabcolsep}{4pt}\renewcommand{\arraystretch}{1.12}
\begin{tabularx}{\linewidth}{@{}>{\hsize=0.66102\hsize\linewidth=\hsize\raggedright\arraybackslash}X>{\hsize=1.16949\hsize\linewidth=\hsize\raggedright\arraybackslash}X>{\hsize=1.16949\hsize\linewidth=\hsize\raggedright\arraybackslash}X@{}}
\toprule
\textbf{Aspect} & \textbf{CoverMem full-frame memory} & \textbf{Structured spatial/object memory} \\
\midrule
Answer evidence & Source appearance, details, and surrounding context & Explicit entities, geometry, or relations \\
Inputs & Association, feature, and depth-based quality preprocessing & Its reconstruction, fusion, and identity pipeline \\
Coverage target & Recorded quality witnesses and representative views & Targets defined by its representation and updates \\
Relations & Implicit in images and requirement matching & Explicitly queryable when represented \\
Error sources & Association and perception errors & Association and reconstruction errors \\
\bottomrule
\end{tabularx}
\end{table}

Table~\ref{tab:supp-memory-comparison} summarizes the differences between full-frame and structured spatial/object memories. A full-frame answer interface does not make the whole pipeline geometry-free: depth participates in visibility and preprocessing can use geometry. CoverMem does not construct every object pair or distance. Localization metrics of streaming object memories also differ from QA scores under the present protocol. These distinctions explain the operating setting rather than imply universal dominance over structured or long-video selectors.

\FloatBarrier
\section{Selection traces and qualitative analysis}\label{app:D}
\setcounter{table}{0}\renewcommand{\thetable}{D\arabic{table}}
\setcounter{figure}{0}\renewcommand{\thefigure}{D\arabic{figure}}

\subsection{Worked complementary-selection example}\label{app:trace}

Tables~\ref{tab:supp-worked-input} and~\ref{tab:supp-worked-selection} report the candidate scores and stepwise selection results in a worked example. Consider a question about the color and material of a chair next to a desk, as well as the chair's position relative to a cabinet. Let $d_1,d_2,d_3$ denote these unknown information needs. Frame labels below belong to this worked example rather than source-frame IDs from a main-paper scene.

\begin{table}[t]\centering
\caption{Fixed normalized relevance and support in the worked example.}\label{tab:supp-worked-input}
\small\setlength{\tabcolsep}{4pt}\renewcommand{\arraystretch}{1.12}
\begin{tabularx}{\linewidth}{@{}>{\hsize=1.00000\hsize\linewidth=\hsize\raggedright\arraybackslash}X>{\hsize=1.00000\hsize\linewidth=\hsize\raggedright\arraybackslash}X>{\hsize=1.00000\hsize\linewidth=\hsize\raggedright\arraybackslash}X>{\hsize=1.00000\hsize\linewidth=\hsize\raggedright\arraybackslash}X>{\hsize=1.00000\hsize\linewidth=\hsize\raggedright\arraybackslash}X@{}}
\toprule
\textbf{Candidate} & \textbf{Relevance} & \textbf{$d_1$} & \textbf{$d_2$} & \textbf{$d_3$} \\
\midrule
fA & 1 & 0.95 & 0.30 & 0.15 \\
fB & 0.75 & 0.90 & 0.35 & 0.20 \\
fC & 0.66 & 0.30 & 0.92 & 0.75 \\
fD & 0.65 & 0.10 & 0.20 & 0.94 \\
fE & 0 & 0 & 0 & 0 \\
\bottomrule
\end{tabularx}
\end{table}

\begin{table}[t]\centering
\caption{Ranking with $\alpha=0.5$, $\lambda=0.1$, $\tau=0.05$, and $K=3$; gain is additional support and red. visual redundancy.}\label{tab:supp-worked-selection}
\small\setlength{\tabcolsep}{4pt}\renewcommand{\arraystretch}{1.12}
\begin{tabularx}{\linewidth}{@{}>{\hsize=1.00000\hsize\linewidth=\hsize\raggedright\arraybackslash}X>{\hsize=1.00000\hsize\linewidth=\hsize\raggedright\arraybackslash}X>{\hsize=1.00000\hsize\linewidth=\hsize\raggedright\arraybackslash}X>{\hsize=1.00000\hsize\linewidth=\hsize\raggedright\arraybackslash}X>{\hsize=1.00000\hsize\linewidth=\hsize\raggedright\arraybackslash}X>{\hsize=1.00000\hsize\linewidth=\hsize\raggedright\arraybackslash}X@{}}
\toprule
\textbf{Step} & \textbf{Frame} & \textbf{Gain} & \textbf{Red.} & \textbf{Score} & \textbf{Pick} \\
\midrule
1 & fA & 0.4667 & 0.00 & 0.7333 & Yes \\
1 & fB & 0.4833 & 0.00 & 0.6167 &  \\
1 & fC & 0.6567 & 0.00 & 0.6583 &  \\
1 & fD & 0.4133 & 0.00 & 0.5317 &  \\
1 & fE & 0.0000 & 0.00 & 0.0000 &  \\
2 & fB & 0.0333 & 1.00 & 0.2917 &  \\
2 & fC & 0.4067 & 0.30 & 0.5033 & Yes \\
2 & fD & 0.2633 & 0.00 & 0.4567 &  \\
2 & fE & 0.0000 & 0.20 & -0.0200 &  \\
3 & fB & 0.0000 & 1.00 & 0.2750 &  \\
3 & fD & 0.0633 & 0.35 & 0.3217 & Yes \\
3 & fE & 0.0000 & 0.20 & -0.0200 &  \\
\bottomrule
\end{tabularx}
\end{table}

Frame fA is selected first. Frame fB then adds little support despite relevance, whereas fC adds material and relation evidence. Frame fD next strengthens the remaining relation need. The budget ends selection with best supports $(0.95,0.92,0.94)$ and mean 0.9367. Relevance alone selects fA, fB, fC, whose mean best support is 0.8733. This illustrates the marginal update without asserting complete instance coverage for every broad list requirement.

\subsection{Existing success and failure cases}\label{app:qualitative}

The four qualitative cases retain the original
method labels, selected frames, complete responses,
and scores. The analyses below examine the displayed
evidence and answer content, highlighting both
successful complementary selection and remaining
limitations in spatial reasoning.

\noindent\textbf{Spatial-relation diagnostics.} Global visual redundancy is a view-level proxy, while requirement support carries semantic matching. Visually similar images can add relation details and visually different images can repeat the same relation. Spatial errors include missing a target or reference, lacking joint relation evidence, and misinterpreting visible evidence.
Spatial understanding improves from 49.12 to 71.18 in the main paper but remains the lowest-scoring category in absolute terms, whereas state recognition shows the smallest improvement. Relation-aware redundancy and pair-specific support are possible extensions rather than components of the submitted method.

\begin{figure*}[p]\centering
\includegraphics[width=\textwidth,height=0.97\textheight,keepaspectratio]{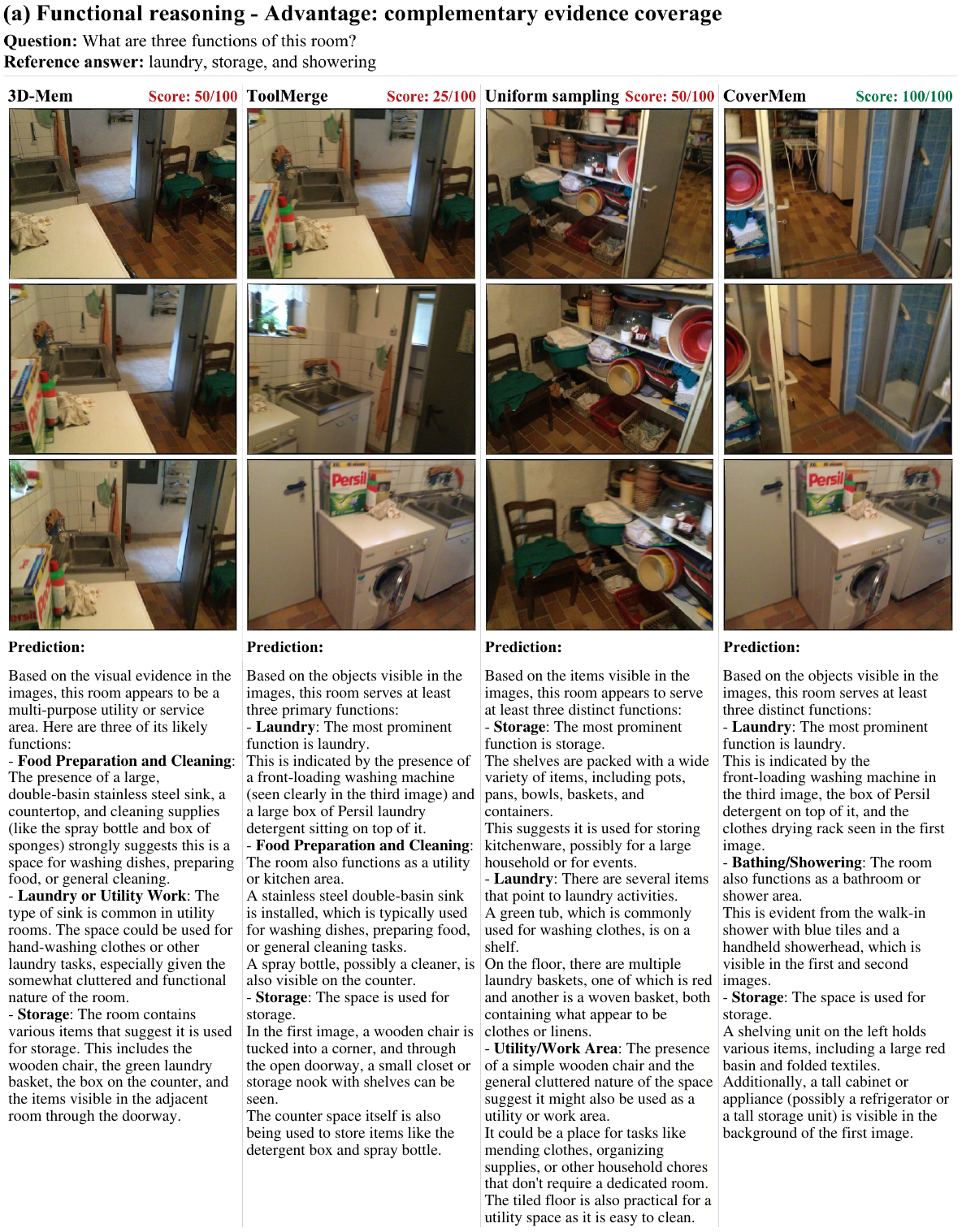}
\caption{\textbf{Functional reasoning.} Selected frames, complete responses, and original scores.}\label{fig:app-case-1}
\end{figure*}

\noindent\textbf{Functional reasoning (Figure~\ref{fig:app-case-1}).} 3D-Mem and ToolMerge identify laundry and storage but substitute food preparation or cleaning for showering. Uniform sampling also recognizes storage and laundry, adding a utility/work-area interpretation. CoverMem combines washing-machine, blue-tiled shower, and shelving views and names all three reference functions. The explicit shower view supplies evidence missing from the other selected sets. This case illustrates the value of allocating the same three-frame budget across different functional regions when the question requests multiple functions.

\begin{figure*}[p]\centering
\includegraphics[width=\textwidth,height=0.97\textheight,keepaspectratio]{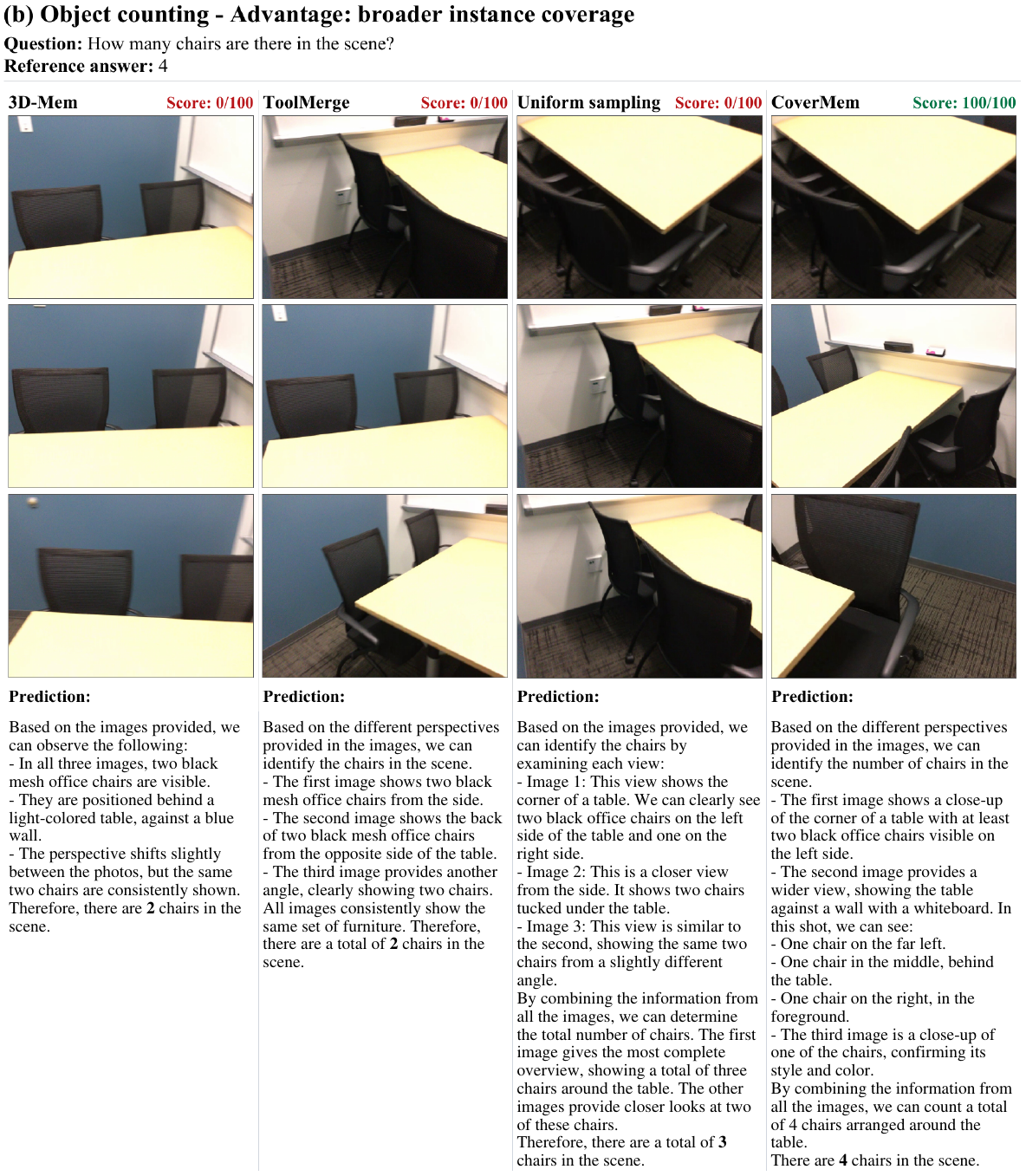}
\caption{\textbf{Object counting.} Selected frames, complete responses, and original scores.}\label{fig:app-case-2}
\end{figure*}

\noindent\textbf{Object counting (Figure~\ref{fig:app-case-2}).} 3D-Mem and ToolMerge report two chairs from partial table views, whereas Uniform sampling reports three. CoverMem includes a wider view from another side of the table and predicts the reference count of four. Compared with the more repetitive views selected by the baselines, its selected frames provide broader visibility of the seating arrangement around the table. However, overlapping viewpoints make it difficult to unambiguously distinguish all chair instances from the displayed evidence alone. This case is consistent with the potential benefit of complementary viewpoints for scene-level counting, although the displayed frames do not establish unambiguous cross-view instance identification.

\begin{figure*}[p]\centering
\includegraphics[width=\textwidth,height=0.97\textheight,keepaspectratio]{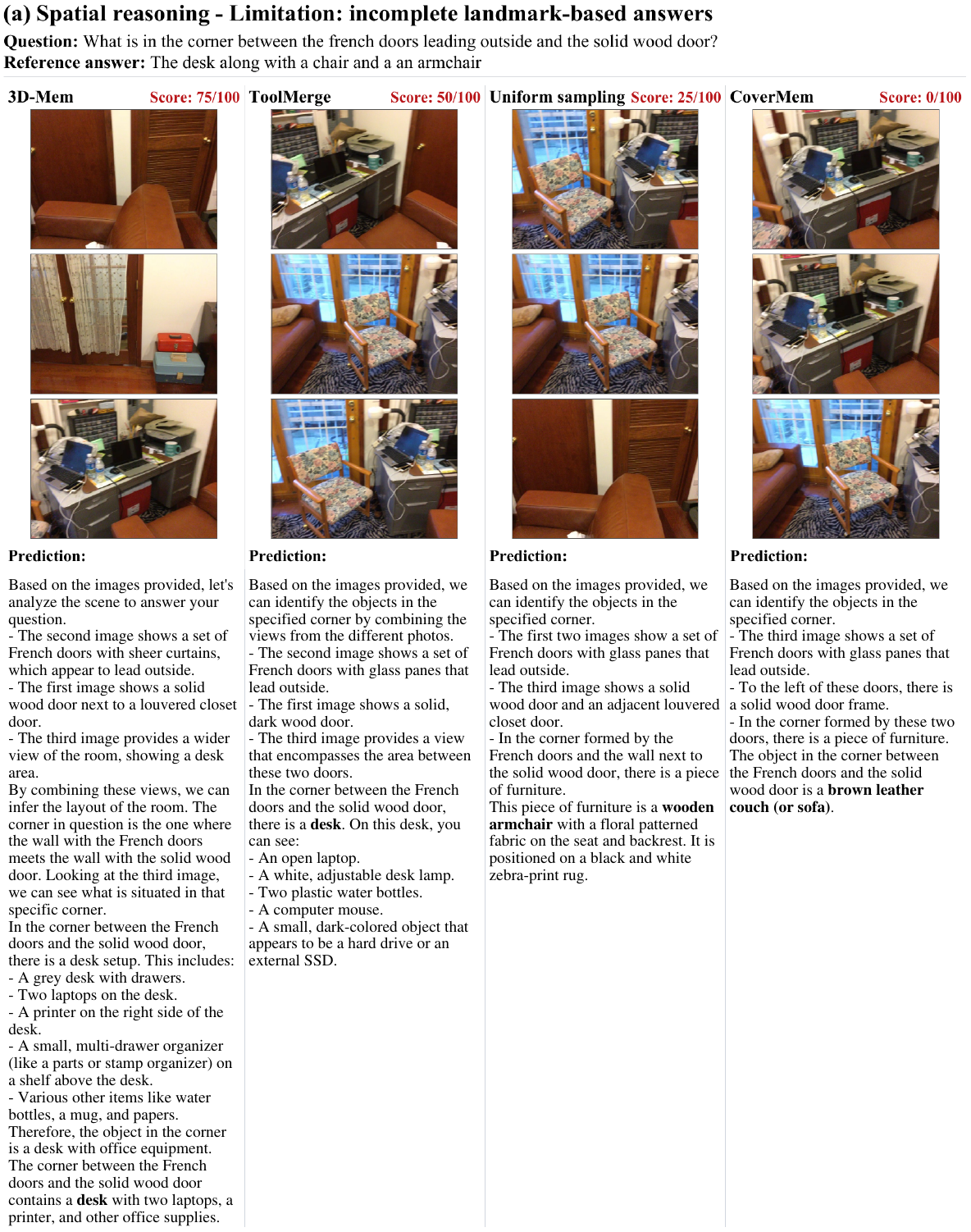}
\caption{\textbf{Landmark binding and answer completeness.} Selected frames, complete responses, and original scores.}\label{fig:app-case-4}
\end{figure*}

\noindent\textbf{Landmark binding and answer completeness (Figure~\ref{fig:app-case-4}).} The reference requests a desk, a chair, and an armchair in a corner defined by two doors. 3D-Mem and ToolMerge identify the desk but elaborate on desktop equipment instead of returning the complete furniture set. Uniform sampling identifies the armchair alone. CoverMem answers couch even though its selected views include the desk area and French doors.
This failure may arise from both incomplete
landmark evidence and incorrect spatial binding
during answer generation.
Although relevant furniture is visible,
the selected frames do not unambiguously
establish the spatial layout defined by the
two doors, leading to an incorrect and
incomplete answer.

\begin{figure*}[p]\centering
\includegraphics[width=\textwidth,height=0.97\textheight,keepaspectratio]{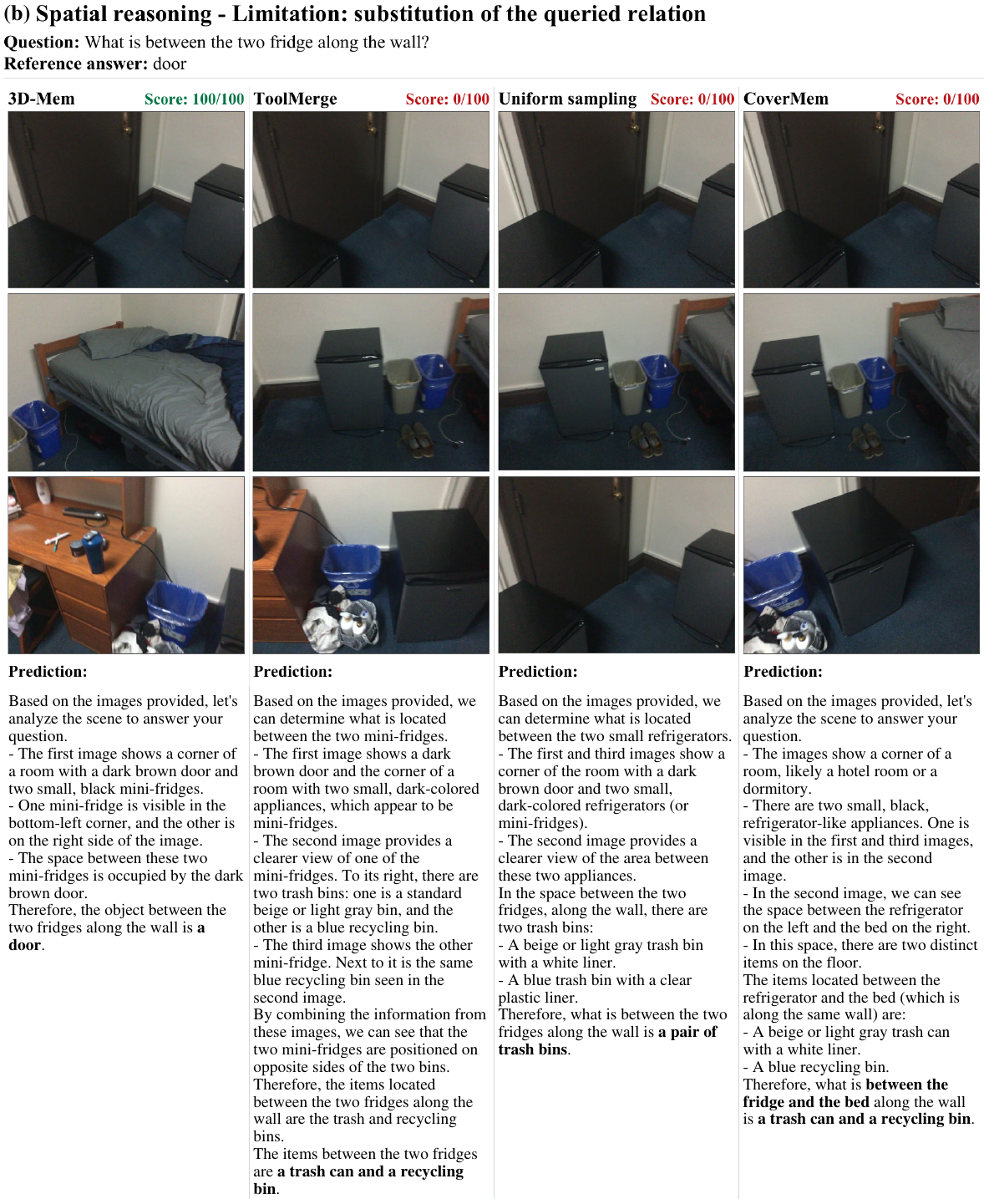}
\caption{
\textbf{Substitution of the queried relation.}
Selected frames, complete responses,
and original scores.
}\label{fig:app-case-6}
\end{figure*}

\noindent\textbf{Queried-relation substitution (Figure~\ref{fig:app-case-6}).} All four selected sets contain a view of the door between the two mini-fridges. 3D-Mem correctly answers door, whereas ToolMerge and uniform sampling incorrectly identify the trash bins as lying between the two fridges.
CoverMem instead changes the queried relation from ``between the two refrigerators'' to ``between a refrigerator and the bed''
before reporting the bins.
The error is directly visible in the response, which describes a different spatial relationship from the one specified in the question. This case separates evidence availability from preserving the target-reference relationship when combining views and generating an answer.

\FloatBarrier

\raggedbottom

\FloatBarrier
\section{Limitations}\label{app:E}
\setcounter{table}{0}\renewcommand{\thetable}{E\arabic{table}}
\setcounter{figure}{0}\renewcommand{\thefigure}{E\arabic{figure}}

\noindent\textbf{Complex requirements and relational evidence.} CoverMem retains reusable visual observations through object-quality and representative-scene coverage, and allocates a limited input budget at question time. For questions involving multiple same-category instances, fine-grained attributes, or complex spatial relations, per-requirement maximum support may still inadequately represent joint evidence. Spatial understanding remains the lowest-scoring
category in the main paper, and the remaining
failure cases show that the answering model can
produce incorrect landmark-conditioned answers
or substitute queried spatial relations even
when relevant visual context is retained. Finer instance-level support and object-pair relation modeling are promising directions.

\noindent\textbf{Upstream information errors.} Construction and selection use preprocessed object associations, region descriptions, and question requirements. Incorrect associations can change coverage targets, while omissions in descriptions or requirements can alter support estimates. Global and regional matching provide visual cues at different scales, and the whole-question fallback handles structurally invalid parser outputs; semantically incorrect but valid requirements require more detailed consistency checking. Incorporating association and description uncertainty into coverage and support estimation may reduce error propagation.

\noindent\textbf{Long-term memory growth.} Local maintenance reduces the frames reconsidered per update, but continually arriving visual content can still increase scene groups and coverage constraints. Fixed representatives and coverage preservation prioritize retaining recorded information; they do not explicitly address stale information or task-dependent forgetting over long operation. Longer streams motivate hierarchical memory, stale-information updates, and resource-constrained coverage management.

\noindent\textbf{Preprocessing and deployment cost.} CoverMem's compact answer inputs depend on detection, association, quality computation, and feature extraction. The main paper reports 44.29 seconds of construction per history, so deployment benefit depends on memory reuse and observation-update frequency. Question-independent processing can be reused across multiple queries over one history, whereas preprocessing remains a consideration for few-query settings or rapidly changing observation streams. Incremental feature updates and on-demand regional processing could further reduce this cost.

%% file: main.bbl
\begin{thebibliography}{8}
\providecommand{\natexlab}[1]{#1}
\providecommand{\url}[1]{\texttt{#1}}
\expandafter\ifx\csname urlstyle\endcsname\relax
  \providecommand{\doi}[1]{doi: #1}\else
  \providecommand{\doi}{doi: \begingroup \urlstyle{rm}\Url}\fi

\bibitem[Bach(2013)]{bach2013learning}
Francis Bach.
\newblock \emph{Learning with submodular functions: A convex optimization perspective}.
\newblock Now Publishers, 2013.

\bibitem[Chvatal(1979)]{chvatal1979greedy}
Vasek Chvatal.
\newblock A greedy heuristic for the set-covering problem.
\newblock \emph{Mathematics of operations research}, 4\penalty0 (3):\penalty0 233--235, 1979.

\bibitem[Dai et~al.(2017)Dai, Chang, Savva, Halber, Funkhouser, and Nie{\ss}ner]{dai2017scannet}
Angela Dai, Angel~X Chang, Manolis Savva, Maciej Halber, Thomas Funkhouser, and Matthias Nie{\ss}ner.
\newblock Scannet: Richly-annotated 3d reconstructions of indoor scenes.
\newblock In \emph{2017 IEEE conference on computer vision and pattern recognition (CVPR)}, pages 2432--2443. IEEE, 2017.

\bibitem[Gu et~al.(2024)Gu, Kuwajerwala, Morin, Jatavallabhula, Sen, Agarwal, Rivera, Paul, Ellis, Chellappa, et~al.]{gu2024conceptgraphs}
Qiao Gu, Ali Kuwajerwala, Sacha Morin, Krishna~Murthy Jatavallabhula, Bipasha Sen, Aditya Agarwal, Corban Rivera, William Paul, Kirsty Ellis, Rama Chellappa, et~al.
\newblock Conceptgraphs: Open-vocabulary 3d scene graphs for perception and planning.
\newblock In \emph{2024 IEEE International Conference on Robotics and Automation (ICRA)}, pages 5021--5028. IEEE, 2024.

\bibitem[Majumdar et~al.(2024)Majumdar, Ajay, Zhang, Putta, Yenamandra, Henaff, Silwal, Mcvay, Maksymets, Arnaud, et~al.]{majumdar2024openeqa}
Arjun Majumdar, Anurag Ajay, Xiaohan Zhang, Pranav Putta, Sriram Yenamandra, Mikael Henaff, Sneha Silwal, Paul Mcvay, Oleksandr Maksymets, Sergio Arnaud, et~al.
\newblock Openeqa: Embodied question answering in the era of foundation models.
\newblock In \emph{2024 IEEE/CVF Conference on Computer Vision and Pattern Recognition (CVPR)}, pages 16488--16498. IEEE, 2024.

\bibitem[Nemhauser et~al.(1978)Nemhauser, Wolsey, and Fisher]{nemhauser1978analysis}
George~L Nemhauser, Laurence~A Wolsey, and Marshall~L Fisher.
\newblock An analysis of approximations for maximizing submodular set functions---i.
\newblock \emph{Mathematical programming}, 14\penalty0 (1):\penalty0 265--294, 1978.

\bibitem[Wolsey(1982)]{wolsey1982analysis}
Laurence~A Wolsey.
\newblock An analysis of the greedy algorithm for the submodular set covering problem.
\newblock \emph{Combinatorica}, 2\penalty0 (4):\penalty0 385--393, 1982.

\bibitem[Yang et~al.(2025)Yang, Yang, Zhou, Chen, Zhang, Du, and Gan]{yang20253d}
Yuncong Yang, Han Yang, Jiachen Zhou, Peihao Chen, Hongxin Zhang, Yilun Du, and Chuang Gan.
\newblock 3d-mem: 3d scene memory for embodied exploration and reasoning.
\newblock In \emph{2025 IEEE/CVF Conference on Computer Vision and Pattern Recognition (CVPR)}, pages 17294--17303. IEEE, 2025.

\end{thebibliography}


\begin{thebibliography}{42}
\providecommand{\natexlab}[1]{#1}
\providecommand{\url}[1]{\texttt{#1}}
\expandafter\ifx\csname urlstyle\endcsname\relax
  \providecommand{\doi}[1]{doi: #1}\else
  \providecommand{\doi}{doi: \begingroup \urlstyle{rm}\Url}\fi

\bibitem[Ali et~al.(2025)Ali, Nair, Wong, Cui, and Chen]{ali2025graphpad}
Muhammad~Qasim Ali, Saeejith Nair, Alexander Wong, Yuchen Cui, and Yuhao Chen.
\newblock Graphpad: Inference-time 3d scene graph updates for embodied question answering.
\newblock \emph{arXiv preprint arXiv:2506.01174}, 2025.

\bibitem[Anwar et~al.(2025)Anwar, Welsh, Biswas, Pouya, and Chang]{anwar2025remembr}
Abrar Anwar, John Welsh, Joydeep Biswas, Soha Pouya, and Yan Chang.
\newblock Remembr: Building and reasoning over long-horizon spatio-temporal memory for robot navigation.
\newblock In \emph{2025 IEEE International Conference on Robotics and Automation (ICRA)}, pages 2838--2845. IEEE, 2025.

\bibitem[Azuma et~al.(2022)Azuma, Miyanishi, Kurita, and Kawanabe]{azuma2022scanqa}
Daichi Azuma, Taiki Miyanishi, Shuhei Kurita, and Motoaki Kawanabe.
\newblock Scanqa: 3d question answering for spatial scene understanding.
\newblock In \emph{2022 IEEE/CVF Conference on Computer Vision and Pattern Recognition (CVPR)}, pages 19107--19117. IEEE, 2022.

\bibitem[Bach(2013)]{bach2013learning}
Francis Bach.
\newblock \emph{Learning with submodular functions: A convex optimization perspective}.
\newblock Now Publishers, 2013.

\bibitem[Carbonell and Goldstein(1998)]{carbonell1998use}
Jaime~G Carbonell and Jade Goldstein.
\newblock The use of mmr, diversity-based reranking for reordering documents and producing summaries.
\newblock In \emph{SIGIR}, pages 290941--291025, 1998.

\bibitem[Chen et~al.(2026)Chen, Zeng, Luo, Xie, Lin, Ji, Zhang, and Zheng]{chen2026wavelet}
Wang Chen, Yuhui Zeng, Yongdong Luo, Tianyu Xie, Luojun Lin, Jiayi Ji, Yan Zhang, and Xiawu Zheng.
\newblock Wavelet-based frame selection by detecting semantic boundary for long video understanding.
\newblock \emph{arXiv preprint arXiv:2603.00512}, 2026.

\bibitem[Cheng et~al.(2025)Cheng, Li, Sun, Min, Bedi, and Bera]{cheng2025efficienteqa}
Kai Cheng, Zhengyuan Li, Xingpeng Sun, Byung-Cheol Min, Amrit~Singh Bedi, and Aniket Bera.
\newblock Efficienteqa: An efficient approach to open-vocabulary embodied question answering.
\newblock In \emph{2025 IEEE/RSJ International Conference on Intelligent Robots and Systems (IROS)}, pages 20357--20364. IEEE, 2025.

\bibitem[Chvatal(1979)]{chvatal1979greedy}
Vasek Chvatal.
\newblock A greedy heuristic for the set-covering problem.
\newblock \emph{Mathematics of operations research}, 4\penalty0 (3):\penalty0 233--235, 1979.

\bibitem[Dai et~al.(2017)Dai, Chang, Savva, Halber, Funkhouser, and Nie{\ss}ner]{dai2017scannet}
Angela Dai, Angel~X Chang, Manolis Savva, Maciej Halber, Thomas Funkhouser, and Matthias Nie{\ss}ner.
\newblock Scannet: Richly-annotated 3d reconstructions of indoor scenes.
\newblock In \emph{2017 IEEE conference on computer vision and pattern recognition (CVPR)}, pages 2432--2443. IEEE, 2017.

\bibitem[Das et~al.(2018)Das, Datta, Gkioxari, Lee, Parikh, and Batra]{das2018embodied}
Abhishek Das, Samyak Datta, Georgia Gkioxari, Stefan Lee, Devi Parikh, and Dhruv Batra.
\newblock Embodied question answering.
\newblock In \emph{2018 IEEE/CVF Conference on Computer Vision and Pattern Recognition Workshops (CVPRW)}, pages 2135--213509. IEEE, 2018.

\bibitem[Fan et~al.(2025)Fan, Ma, Su, Guo, Wu, Chen, and Li]{fan2025embodied}
Yue Fan, Xiaojian Ma, Rongpeng Su, Jun Guo, Rujie Wu, Xi Chen, and Qing Li.
\newblock Embodied videoagent: Persistent memory from egocentric videos and embodied sensors enables dynamic scene understanding.
\newblock In \emph{2025 IEEE/CVF International Conference on Computer Vision (ICCV)}, pages 6342--6352. IEEE, 2025.

\bibitem[Ginting et~al.(2025)Ginting, Kim, Meng, Reinke, Krishna, Kayhani, Peltzer, Fan, Shaban, Kim, et~al.]{ginting2025enter}
Muhammad~Fadhil Ginting, Dong-Ki Kim, Xiangyun Meng, Andrzej Reinke, Bandi~Jai Krishna, Navid Kayhani, Oriana Peltzer, David~D Fan, Amirreza Shaban, Sung-Kyun Kim, et~al.
\newblock Enter the mind palace: Reasoning and planning for long-term active embodied question answering.
\newblock \emph{arXiv preprint arXiv:2507.12846}, 2025.

\bibitem[Gordon et~al.(2018)Gordon, Kembhavi, Rastegari, Redmon, Fox, and Farhadi]{gordon2018iqa}
Daniel Gordon, Aniruddha Kembhavi, Mohammad Rastegari, Joseph Redmon, Dieter Fox, and Ali Farhadi.
\newblock Iqa: Visual question answering in interactive environments.
\newblock In \emph{2018 IEEE/CVF Conference on Computer Vision and Pattern Recognition}, pages 4089--4098. IEEE, 2018.

\bibitem[Gu et~al.(2024)Gu, Kuwajerwala, Morin, Jatavallabhula, Sen, Agarwal, Rivera, Paul, Ellis, Chellappa, et~al.]{gu2024conceptgraphs}
Qiao Gu, Ali Kuwajerwala, Sacha Morin, Krishna~Murthy Jatavallabhula, Bipasha Sen, Aditya Agarwal, Corban Rivera, William Paul, Kirsty Ellis, Rama Chellappa, et~al.
\newblock Conceptgraphs: Open-vocabulary 3d scene graphs for perception and planning.
\newblock In \emph{2024 IEEE International Conference on Robotics and Automation (ICRA)}, pages 5021--5028. IEEE, 2024.

\bibitem[Gupta et~al.(2017)Gupta, Krishnaswamy, Kumar, and Panigrahi]{gupta2017online}
Anupam Gupta, Ravishankar Krishnaswamy, Amit Kumar, and Debmalya Panigrahi.
\newblock Online and dynamic algorithms for set cover.
\newblock In \emph{Proceedings of the 49th Annual ACM SIGACT Symposium on Theory of Computing}, pages 537--550, 2017.

\bibitem[Hong et~al.(2023)Hong, Zhen, Chen, Zheng, Du, Chen, and Gan]{hong20233d}
Yining Hong, Haoyu Zhen, Peihao Chen, Shuhong Zheng, Yilun Du, Zhenfang Chen, and Chuang Gan.
\newblock 3d-llm: Injecting the 3d world into large language models.
\newblock \emph{Advances in Neural Information Processing Systems}, 36:\penalty0 20482--20494, 2023.

\bibitem[Huang et~al.(2023{\natexlab{a}})Huang, Mees, Zeng, and Burgard]{huang2023visual}
Chenguang Huang, Oier Mees, Andy Zeng, and Wolfram Burgard.
\newblock Visual language maps for robot navigation.
\newblock In \emph{2023 IEEE International Conference on Robotics and Automation (ICRA)}, pages 10608--10615. IEEE, 2023{\natexlab{a}}.

\bibitem[Huang et~al.(2026)Huang, Chen, Tsao, Lee, Su, Xia, Peng, Hwang, Sun, and Kuo]{huang2026memory}
Hsiang-Wei Huang, Fu-Chen Chen, Li-Wu Tsao, Cheng-Han Lee, Che-Chun Su, Lu Xia, Ronghui Peng, Jenq-Neng Hwang, Min Sun, and Cheng-Hao Kuo.
\newblock Memory tree guided key frame querying for efficient 3d question answering.
\newblock In \emph{European Conference on Computer Vision}, pages 337--354. Springer, 2026.

\bibitem[Huang et~al.(2023{\natexlab{b}})Huang, Yong, Ma, Linghu, Li, Wang, Li, Zhu, Jia, and Huang]{huang2023embodied}
Jiangyong Huang, Silong Yong, Xiaojian Ma, Xiongkun Linghu, Puhao Li, Yan Wang, Qing Li, Song-Chun Zhu, Baoxiong Jia, and Siyuan Huang.
\newblock An embodied generalist agent in 3d world.
\newblock \emph{arXiv preprint arXiv:2311.12871}, 2023{\natexlab{b}}.

\bibitem[Jatavallabhula et~al.(2023)Jatavallabhula, Kuwajerwala, Gu, Omama, Chen, Maalouf, Li, Iyer, Saryazdi, Keetha, et~al.]{jatavallabhula2023conceptfusion}
Krishna~Murthy Jatavallabhula, Alihusein Kuwajerwala, Qiao Gu, Mohd Omama, Tao Chen, Alaa Maalouf, Shuang Li, Ganesh Iyer, Soroush Saryazdi, Nikhil Keetha, et~al.
\newblock Conceptfusion: Open-set multimodal 3d mapping.
\newblock \emph{arXiv preprint arXiv:2302.07241}, 2023.

\bibitem[Jiang et~al.(2025)Jiang, Liu, Chen, Luo, Chen, Pan, Li, and Lin]{jiang2025beyond}
Kaixuan Jiang, Yang Liu, Weixing Chen, Jingzhou Luo, Ziliang Chen, Ling Pan, Guanbin Li, and Liang Lin.
\newblock Beyond the destination: A novel benchmark for exploration-aware embodied question answering.
\newblock In \emph{2025 IEEE/CVF International Conference on Computer Vision (ICCV)}, pages 9091--9101. IEEE, 2025.

\bibitem[Kerbl et~al.(2023)Kerbl, Kopanas, Leimk{\"u}hler, Drettakis, et~al.]{kerbl20233d}
Bernhard Kerbl, Georgios Kopanas, Thomas Leimk{\"u}hler, George Drettakis, et~al.
\newblock 3d gaussian splatting for real-time radiance field rendering.
\newblock \emph{ACM Trans. Graph.}, 42\penalty0 (4):\penalty0 139--1, 2023.

\bibitem[Kerr et~al.(2023)Kerr, Kim, Goldberg, Kanazawa, and Tancik]{kerr2023lerf}
Justin Kerr, Chung~Min Kim, Ken Goldberg, Angjoo Kanazawa, and Matthew Tancik.
\newblock Lerf: Language embedded radiance fields.
\newblock In \emph{2023 IEEE/CVF International Conference on Computer Vision (ICCV)}, pages 19672--19682. IEEE, 2023.

\bibitem[Kitamura and Kanezaki(2026)]{kitamura2026informative}
Kaname Kitamura and Asako Kanezaki.
\newblock Informative viewpoint selection for episodic-memory embodied question answering using omnidirectional images.
\newblock \emph{arXiv preprint arXiv:2609.33288}, 2026.

\bibitem[Lu et~al.(2026)Lu, Du, Liu, Zhou, Wang, and Yin]{lu2026gsmem}
Yiren Lu, Yi Du, Disheng Liu, Yunlai Zhou, Chen Wang, and Yu Yin.
\newblock Gsmem: 3d gaussian splatting as persistent spatial memory for zero-shot embodied exploration and reasoning.
\newblock \emph{arXiv preprint arXiv:2603.19137}, 2026.

\bibitem[Majumdar et~al.(2024)Majumdar, Ajay, Zhang, Putta, Yenamandra, Henaff, Silwal, Mcvay, Maksymets, Arnaud, et~al.]{majumdar2024openeqa}
Arjun Majumdar, Anurag Ajay, Xiaohan Zhang, Pranav Putta, Sriram Yenamandra, Mikael Henaff, Sneha Silwal, Paul Mcvay, Oleksandr Maksymets, Sergio Arnaud, et~al.
\newblock Openeqa: Embodied question answering in the era of foundation models.
\newblock In \emph{2024 IEEE/CVF Conference on Computer Vision and Pattern Recognition (CVPR)}, pages 16488--16498. IEEE, 2024.

\bibitem[Nemhauser et~al.(1978)Nemhauser, Wolsey, and Fisher]{nemhauser1978analysis}
George~L Nemhauser, Laurence~A Wolsey, and Marshall~L Fisher.
\newblock An analysis of approximations for maximizing submodular set functions---i.
\newblock \emph{Mathematical programming}, 14\penalty0 (1):\penalty0 265--294, 1978.

\bibitem[Ong and Jang(2025)]{ong2025r}
Hyobin Ong and Minsu Jang.
\newblock R-eqa: Retrieval-augmented generation for embodied question answering.
\newblock In \emph{Proc. IEEE/CVF Conf. Comput. Vis. Pattern Recognit. Workshops (CVPRW)}, pages 1--3, 2025.

\bibitem[Peng et~al.(2026)Peng, Song, Li, Li, Zhou, Ji, and Tian]{peng2026qca}
Jun Peng, Baiyang Song, Jie Li, Hui Li, Yiyi Zhou, Rongrong Ji, and Yonghong Tian.
\newblock Qca: Query-and content-aware keyframe selection for long video understanding.
\newblock In \emph{European Conference on Computer Vision}, pages 18--35. Springer, 2026.

\bibitem[Peng et~al.(2023)Peng, Genova, Jiang, Tagliasacchi, Pollefeys, and Funkhouser]{peng2023openscene}
Songyou Peng, Kyle Genova, Chiyu Jiang, Andrea Tagliasacchi, Marc Pollefeys, and Thomas Funkhouser.
\newblock Openscene: 3d scene understanding with open vocabularies.
\newblock In \emph{2023 IEEE/CVF Conference on Computer Vision and Pattern Recognition (CVPR)}, pages 815--824. IEEE, 2023.

\bibitem[Radford(2021)]{radford2021learning}
A Radford.
\newblock Learning transferable visual models from natural language supervision.
\newblock \emph{arXiv preprint arXiv:2103.00020}, 2021.

\bibitem[Ren et~al.(2024)Ren, Clark, Dixit, Itkina, Majumdar, and Sadigh]{ren2024explore}
Allen~Z Ren, Jaden Clark, Anushri Dixit, Masha Itkina, Anirudha Majumdar, and Dorsa Sadigh.
\newblock Explore until confident: Efficient exploration for embodied question answering.
\newblock \emph{arXiv preprint arXiv:2403.15941}, 2024.

\bibitem[Saxena et~al.(2024)Saxena, Buchanan, Paxton, Liu, Chen, Vaskevicius, Palmieri, Francis, and Kroemer]{saxena2024grapheqa}
Saumya Saxena, Blake Buchanan, Chris Paxton, Peiqi Liu, Bingqing Chen, Narunas Vaskevicius, Luigi Palmieri, Jonathan Francis, and Oliver Kroemer.
\newblock Grapheqa: Using 3d semantic scene graphs for real-time embodied question answering.
\newblock \emph{arXiv preprint arXiv:2412.14480}, 2024.

\bibitem[Shlapentokh-Rothman et~al.(2026)Shlapentokh-Rothman, Garg, Wang, and Hoiem]{shlapentokh2026decomposing}
Michal Shlapentokh-Rothman, Prachi Garg, Yu-Xiong Wang, and Derek Hoiem.
\newblock Decomposing queries into tool calls for long-video keyframe retrieval.
\newblock \emph{arXiv preprint arXiv:2605.23826}, 2026.

\bibitem[Sohn et~al.(2026)Sohn, Dillitzer, Corso, and Sax]{sohn2026r4}
Tin~Stribor Sohn, Maximilian Dillitzer, Jason~J Corso, and Eric Sax.
\newblock R4: Retrieval-augmented reasoning for vision-language models in 4d spatio-temporal space.
\newblock In \emph{Proceedings of the IEEE/CVF Conference on Computer Vision and Pattern Recognition}, pages 38711--38721, 2026.

\bibitem[Wang et~al.(2024)Wang, Zhang, Zohar, and Yeung-Levy]{wang2024videoagent}
Xiaohan Wang, Yuhui Zhang, Orr Zohar, and Serena Yeung-Levy.
\newblock Videoagent: Long-form video understanding with large language model as agent.
\newblock In \emph{European Conference on Computer Vision}, pages 58--76. Springer, 2024.

\bibitem[Wang et~al.(2025)Wang, Yu, Stengel-Eskin, Yoon, Cheng, Bertasius, and Bansal]{wang2025videotree}
Ziyang Wang, Shoubin Yu, Elias Stengel-Eskin, Jaehong Yoon, Feng Cheng, Gedas Bertasius, and Mohit Bansal.
\newblock Videotree: Adaptive tree-based video representation for llm reasoning on long videos.
\newblock In \emph{2025 IEEE/CVF Conference on Computer Vision and Pattern Recognition (CVPR)}, pages 3272--3282. IEEE, 2025.

\bibitem[Wolsey(1982)]{wolsey1982analysis}
Laurence~A Wolsey.
\newblock An analysis of the greedy algorithm for the submodular set covering problem.
\newblock \emph{Combinatorica}, 2\penalty0 (4):\penalty0 385--393, 1982.

\bibitem[Xu et~al.(2024)Xu, Huang, Wang, Chen, Pang, and Lin]{xu2024vlm}
Runsen Xu, Zhiwei Huang, Tai Wang, Yilun Chen, Jiangmiao Pang, and Dahua Lin.
\newblock Vlm-grounder: A vlm agent for zero-shot 3d visual grounding.
\newblock \emph{arXiv preprint arXiv:2410.13860}, 2024.

\bibitem[Yang et~al.(2025)Yang, Yang, Zhou, Chen, Zhang, Du, and Gan]{yang20253d}
Yuncong Yang, Han Yang, Jiachen Zhou, Peihao Chen, Hongxin Zhang, Yilun Du, and Chuang Gan.
\newblock 3d-mem: 3d scene memory for embodied exploration and reasoning.
\newblock In \emph{2025 IEEE/CVF Conference on Computer Vision and Pattern Recognition (CVPR)}, pages 17294--17303. IEEE, 2025.

\bibitem[Yu-Ji et~al.(2026)Yu-Ji, Lee, Jun-Seong, Hyeon-Woo, Kim, Kwon, Wang, Choe, and Oh]{yu2026splatreasoner}
Kim Yu-Ji, Dahye Lee, Kim Jun-Seong, Nam Hyeon-Woo, GeonU Kim, Yongjin Kwon, Yu-Chiang~Frank Wang, Jaesung Choe, and Tae-Hyun Oh.
\newblock Splatreasoner: Enhancing embodied reasoning and grounding by novel view synthesis.
\newblock In \emph{European Conference on Computer Vision}, pages 133--151. Springer, 2026.

\bibitem[Zhang et~al.(2026)Zhang, Savaliya, Siddiqui, and Sachdeva]{zhang2026fast}
Haochen Zhang, Nirav Savaliya, Faizan Siddiqui, and Enna Sachdeva.
\newblock Fast-eqa: Efficient embodied question answering with global and local region relevancy.
\newblock In \emph{2026 IEEE/CVF Winter Conference on Applications of Computer Vision (WACV)}, pages 1664--1673. IEEE, 2026.

\end{thebibliography}
